\documentclass[10pt,journal,compsoc]{IEEEtran}
\ifCLASSOPTIONcompsoc
  \usepackage[nocompress]{cite}
\else
  \usepackage{cite}
\fi
\ifCLASSINFOpdf
  \usepackage[pdftex]{graphicx}
\else
\fi
\usepackage{amsmath}
\usepackage[ruled]{algorithm2e}

\ifCLASSOPTIONcompsoc
 \usepackage[caption=false,font=footnotesize,labelfont=sf,textfont=sf]{subfig}
\else
 \usepackage[caption=false,font=footnotesize]{subfig}
\fi

\usepackage{url}
\usepackage{booktabs}
\usepackage[normalem]{ulem}
\useunder{\uline}{\ul}{}
\usepackage{multirow}

\usepackage{amssymb}
\usepackage{bm}
\usepackage{bbm}
\usepackage[capitalize]{cleveref}
\crefname{section}{Section}{Secs.}
\Crefname{section}{Section}{Sections}
\Crefname{table}{Table}{Tables}
\crefname{table}{Tab.}{Tabs.}
\crefname{table}{Tab.}{Tabs.}

\newlength{\tabcolseporig}
\begin{document}
% \title{Instance Adaptive Loss Smoothness Enhancement for Improving Adversarial Training}
\title{Improved Adversarial Training Through Adaptive Instance-wise Loss Smoothing}

\author{Lin~Li
        and~Michael~Spratling
        % <-this % stops a space
\IEEEcompsocitemizethanks{
\IEEEcompsocthanksitem Lin Li is with the Department of Informatics, King's College London, 30 Aldwych, London, WC2B 4BG, UK. \protect\\
E-mail: lin.3.li@kcl.ac.uk
\IEEEcompsocthanksitem Michael Spratling is with the Department of Informatics, King's College London, 30 Aldwych, London, WC2B 4BG, UK. \protect\\
E-mail: michael.spratling@kcl.ac.uk}% <-this % stops an unwanted space
\thanks{Manuscript received January 2023}}


\markboth{Journal of \LaTeX\ Class Files,~Vol.~14, No.~8, August~2015}%
{Shell \MakeLowercase{\textit{et al.}}: Bare Demo of IEEEtran.cls for Computer Society Journals}



\IEEEtitleabstractindextext{%
\begin{abstract}
Deep neural networks can be easily fooled into making incorrect predictions through corruption of the input by adversarial perturbations: human-imperceptible artificial noise. So far adversarial training has been the most successful defense against such adversarial attacks.
This work focuses on improving adversarial training to boost adversarial robustness. We first analyze, from an instance-wise perspective, how adversarial vulnerability evolves during adversarial training. We find that during training an overall reduction of adversarial loss is achieved by sacrificing a considerable proportion of training samples to be  more vulnerable to adversarial attack, which results in an uneven distribution of adversarial vulnerability among data. Such "uneven vulnerability", is prevalent across several popular robust training methods and, more importantly, relates to overfitting in adversarial training. Motivated by this observation, we propose a new adversarial training method: Instance-adaptive Smoothness Enhanced Adversarial Training (ISEAT). It jointly smooths both input and weight loss landscapes in an adaptive, instance-specific, way to enhance robustness more for those samples with higher adversarial vulnerability. Extensive experiments demonstrate the superiority of our method over existing defense methods. Noticeably, our method, when combined with the latest data augmentation and semi-supervised learning techniques, achieves state-of-the-art robustness against $\ell_{\infty}$-norm constrained attacks on CIFAR10 of 59.32\% for Wide ResNet34-10 without extra data, and 61.55\% for Wide ResNet28-10 with extra data. Code is available at \url{https://github.com/TreeLLi/Instance-adaptive-Smoothness-Enhanced-AT}.

\end{abstract}

}

\maketitle

\IEEEdisplaynontitleabstractindextext

% \fi
\IEEEpeerreviewmaketitle

\IEEEraisesectionheading{\section{Introduction}\label{sec:introduction}}

\IEEEPARstart{D}{eep} neural networks (DNNs) are well known to be very vulnerable to adversarial attacks \cite{goodfellow_explaining_2015}. Adversarial attacks modify (or “perturb”) natural images (clean examples) to craft adversarial examples in such a way as to fool the target network to make predictions that are dramatically different from those for the corresponding clean examples even when the class of the perturbed input appears unchanged to a human observer. This raises severe security concerns about DNNs, especially as more and more real-world applications are dependent upon such models.

To date, adversarial training (AT) has been found to be the most effective defense against adversarial attacks \cite{athalye_obfuscated_2018}. It is typically formulated as a min-max problem:
\begin{equation}
    \arg \min_{\bm{\theta}} \mathbb{E}[\arg \max_{\bm{\delta}} \mathcal{L}(\bm{x}+\bm{\delta}; \bm{\theta})]
\end{equation}
where the inner maximization searches for the strongest adversarial perturbation $\bm{\delta}$ and the outer optimization searches for the model parameters $\bm{\theta}$ to minimize the loss on the generated adversarial examples. 
One particular limit of vanilla AT \cite{madry_towards_2018} is that all samples in the data set are treated equally during training, neglecting individual differences between samples. Several previous works have made improvements to AT by customizing regularization in an instance-adaptive way. Regularization here can be implemented either  by modifying the method used to generate the training-time adversarial sample or by modifying the strength of regularization applied to the loss function. 
%We defer the discussion of robust regularization to \cref{sec: related works}. 
For example, one popular instance-adaptive strategy is to enhance the strength of the training-time adversarial attack\footnote{The strength of the attack can be modified by scaling the magnitude of the perturbations found by the attack, or by changing the perturbation budget used by the attack, or by changing the number of steps used by the attack.} for hard-to-attack (adversarially benign) samples and/or to diminish the strength of the attack at those easy-to-attack samples \cite{balaji_instance_2019, ding_mma_2020, zhang_attacks_2020, cheng_cat_2022, yang_one_2022}. Other strategies are discussed in detail in \cref{sec: related works}. We extend this line of work to improve AT, but propose a different strategy to distinguish instances (that particularly contrasts with the aforementioned popular method), and propose a new regularizer to jointly smooth both input and weight loss landscapes.

% Although many progresses \cite{wu_adversarial_2020, wang_improving_2020, carmon_unlabeled_2019} have been made to improve adversarial training, the state-of-the-art robustness, typically measured by accuracy against adversarial examples, is still much behind the clean accuracy achieved by standard training, i.e., without adversarial examples used for training. Therefore, adversarial robustness remains an unsolved property and improving adversarial training is important for achieving so.

The proposed approach of improving AT to motivated by an analysis, from an instance-wise perspective, of how adversarial vulnerability (formally defined as in \cref{equ: adversarial vulnerability}) evolves during AT. We first discuss the theoretical possibility that during AT the reduction in the overall adversarial loss may be produced by sacrificing some gradients (allowing them to increase) so that the others can decrease more.
We then demonstrate, empirically, the prevalence of this issue in practice across various datasets and adversarial training methods. We observed that during AT there is an increase in the number of samples that have low- and, more surprisingly, high-vulnerability to adversarial attack. This produces an uneven distribution of robustness among instances, or "uneven vulnerability" for short. Furthermore, we show that for a large proportion of samples with low-vulnerability their margin along the adversarial direction is excessive in the extreme. Specifically, even if modified with a perturbation with a magnitude about 30 times the perturbation budget, such samples can still be correctly classified. We describe such samples as having "disordered robustness". Next, we demonstrate that the uneven vulnerability issue is related to overfitting in AT and can not be mitigated by the popular robust overfitting regularization methods like AWP \cite{wu_adversarial_2020} and RST \cite{carmon_unlabeled_2019}. Therefore, we hypothesize that enforcing an even distribution of adversarial vulnerability among data can improve the robust generalization of AT. Importantly, the proposed method is complimentary to other regularization techniques.

The above insights lead to the proposal of a novel AT approach named Instance-adaptive Smoothness Enhanced Adversarial Training (ISEAT). It jointly enhances both input and weight loss landscape smoothness in an instance-adaptive way. Specifically, it enforces, in addition to standard AT, logit stability against both adversarial input and weight perturbation and the strength of regularization for each instance is adaptive to its adversarial vulnerability. Extensive experiments were conducted to evaluate the performance of the proposed method on various datasets and models. It significantly improves the baseline AT method and outperforms in terms of adversarial robustness previous related methods. Using standard benchmarks the proposed method produces new state-of-the-art robustness of 59.32\% and 61.55\% respectively for settings that do not, and do, use extra data during training. A detailed ablation study was conducted to illuminate the mechanism behind the effectiveness of the proposed method.

\section{Related Works} \label{sec: related works}
Adversarial training is usually categorized as single-step and mutli-step AT according to the number of iterations used for crafting training adversarial examples. The common single-step and multi-step adversaries are Fast Gradient Sign Method (FGSM) \cite{goodfellow_explaining_2015} and Projected Gradient Descent (PGD) \cite{madry_towards_2018}. FGSM uses the sign of the loss gradients w.r.t. the input as the adversarial direction. PGD can be considered as an iterative version of FGSM where the perturbation is projected back onto the $\ell_{\infty}$-norm $\epsilon$-ball at the end of each iteration. For brevity a number is appended to the abbreviation PGD to indicate the number of steps performed when searching for an adversarial image. Hence, for example, PGD50 is used to denote a PGD attack with 50 steps throughout the text. AT is prone to overfitting, which is known as robust overfitting \cite{rice_overfitting_2020}. Specifically, test adversarial robustness degenerates while training adversarial loss declines during the later stage of training. Robust overfitting significantly impairs the performance of AT.

It has been shown previously that adversarial robustness is related to the smoothness of the model's loss landscape w.r.t. the input \cite{simon-gabriel_first-order_2019, moosavi-dezfooli_robustness_2019} and the model weights \cite{wu_adversarial_2020}. Therefore, we summarize existing methods for adversarial robustness in two categories: input loss smoothing and weight loss smoothing. For input loss smoothing, one approach is explicitly regularizing the logits \cite{li_understanding_2023} or the gradients \cite{moosavi-dezfooli_robustness_2019} of each training sample to be the same as any of their neighbor samples within a certain distance. Besides, AT can be concerned as an implicit input loss smoothness regularizer and the strength of regularization is controlled by the direction and the size of perturbation \cite{li_understanding_2023}.
Regarding weight loss smoothing, Adversarial Weight Perturbation (AWP) \cite{wu_adversarial_2020} injects adversarial perturbation into model weights to implicitly smooth weight loss. RWP \cite{yu_robust_2022} found that applying adversarial weight perturbation to only small-loss samples leads to an improved robustness compared to AWP. 
Alternatively, Stochastic Weight Averaging (SWA) \cite{chen_robust_2021} smooths weight by exponentially averaging checkpoints along the training trajectory. Our regularizer combines logit regularization and AWP together to jointly smooth both input and weight loss in a more effective way.

% TODO: most of related works lack a reasoning about why instance adapt so

Many strategies have been proposed to improve AT by regularizing instances unequally. One popular strategy is to adapt the size of the adversarial input perturbation, and so the strength of regularization, to the difficulty of crafting successful adversarial examples. Typically, large perturbations are assigned to hard-to-attack samples in order to produce successful adversarial examples for more effective AT. In tandem, small perturbations are assigned to easy-to-attack samples in order to alleviate over-regularization for a better trade-off between accuracy and robustness. The size of perturbation can be controlled by the number of steps \cite{zhang_attacks_2020}, the perturbation budget \cite{balaji_instance_2019, cheng_cat_2022, yang_one_2022} or an extra scaling factor \cite{ding_mma_2020}. 
Furthermore, this strategy has been also applied to weight loss smoothing in RWP \cite{yu_robust_2022}. 
We argue that this strategy contradicts our finding that hard-to-attack (low-vulnerability in our terms) samples have already been over-regularized so their regularization strength should not be further enlarged. 

The most similar methods to ours are MART \cite{wang_improving_2020} and GAIRAT \cite{zhang_geometry-aware_2021}. MART regularizes KL-divergence between the logit of clean and corresponding adversarial examples, weighted by one minus the prediction confidence on clean examples. Hence, instances with lower clean prediction confidence receive stronger regularization. GAIRAT directly weights the adversarial loss of instances based on their geometric distance to the decision bound, which is measured by the number of iterations required for a successful attack. Instances closer to decision bound (less iterations) are weighted more.
Although there is a connection among prediction confidence, geometric distance and adversarial vulnerability (ours), they are essentially different metrics and the weight schemes based on them thus perform differently. 
Regarding GAIRAT, the computation of geometric distance deeply depends on the training adversary, and hence, severely limits its application, e.g., it can not be applied to single-step AT without using an extra multi-step adversary.
Another similar work is InfoAT \cite{xu_infoat_2022} which like the proposed method  uses logit stability regularization, but weights regularization according to the mutual information of clean examples.

In contrast to the manually crafted strategies described above, LAS-AT \cite{jia_-at_2022} customizes adversarial attack automatically for each instance in a generative adversarial style. The parameters of the attacker, such as the perturbation budget for each instance, are generated on-the-fly by a separate strategy network, which is jointly trained alongside the classification network to maximize adversarial loss, i.e., produce stronger adversarial examples. This approach is more complicated than the alternatives, including ours, and potentially suffers from similar instability issues to GANs \cite{arjovsky_towards_2017}. 

AT can benefit from using more training data to enhance robust generalization. This was theoretically proved for simplified settings like Gaussian models \cite{schmidt_adversarially_2018}. For complicated but realistic settings, training with extra data from either real \cite{carmon_unlabeled_2019}, or synthetic \cite{gowal_improving_2021}, sources dramatically boosts the robust performance of AT and thus becomes a necessary ingredient for achieving state-of-the-art results. However, extra data is usually very expensive or even infeasible to acquire in many tasks, so \cite{li_data_2023} proposes a new data augmentation method, IDBH, specifically designed for AT. Our method is complementary to using extra data and data augmentation and a further boost in robustness is observed when they are combined together (see \cref{sec: benchmarking robustness}).

\section{Uneven Vulnerability and Disordered Robustness} \label{sec: uneven vulnerability}
This section describes two issues of AT: ``uneven vulnerability'' and ``disordered robustness''. We first propose, theoretically, an alternative optimization path for AT which leads to the uneven vulnerability issue. We then demonstrate, empirically, that AT in practice is prone to optimize in this manner, and thus, to producing models that becomes increasingly vulnerable at some samples while robust at some others. We find that the robustness for a considerable proportion of training samples is actually disordered, because the model is insensitive to dramatic perturbations applied to them that should significantly affect the model's output. Last, we relate the uneven vulnerability issue to overfitting in AT and show that robust generalization can be improved by alleviating unevenness.

We adopt a similar notation to that used in \cite{li_understanding_2023}. $\bm{x} \in \mathbb{R}^d$ is a $d$-dimensional sample whose ground truth label is $y$. $\bm{x}_i$ refers to $i$-th sample in dataset $D$ and $x_{i, j}$ refers to the $j$-th dimension of $\bm{x}_i$. Similarly, $x_j$ refers to the $j$-th dimension of an arbitrary sample $\bm{x}$. $D$ is split into two subsets, $D_t$ and $D_e$, for training and testing respectively.
$\bm{\delta}_i \in \mathcal{B}(\bm{x}_i, \epsilon)$ is the adversarial perturbation applied to $\bm{x}_i$ to fool the network. $\mathcal{B}(\bm{x}, \epsilon)$ denotes the $\epsilon$-ball around the example $\bm{x}$ defined for a specific distance measure (the $\ell_{\infty}$-norm in this paper). $\delta_{i, j}$ is the perturbation applied along the dimension $x_{i, j}$, and $\delta_j$ is the perturbation applied along the $j$-th dimension of an arbitrary sample $\bm{x}$. The network is parameterized by $\bm{\theta}$. The output of the network before the final softmax layer (i.e., the logits) is $f(\bm{x};\bm{\theta})$. The class predicted by the network, i.e., the class associated with the highest logit value, is $F(\bm{x};\bm{\theta})$. The predictive loss is $\mathcal{L}(\bm{x}, y; \theta)$ or $\mathcal{L}(\bm{x})$ for short (Cross-Entropy loss was used in all experiments).
The size of a batch of samples used in one step of training is $m$. $M$ denotes the collection of indexes of the samples in a batch.

The experiments in this section were conducted using the default settings with CIFAR10 as defined in \cref{sec: result} unless specified otherwise. The model architecture was PreAct ResNet18 \cite{he_identity_2016}. Robustness was evaluated against 
%the attack Projected Gradient Descent (PGD) \cite{madry_towards_2018} with 50 steps abbreviated as 
PGD50.

\subsection{Adversarial Robustness and Loss Landscape} \label{sec: adversarial formulation}
Adversarial loss can be approximated following \cite{li_understanding_2023} by the sum of clean loss and adversarial vulnerability as
\begin{equation}
    \mathcal{L}(\bm{x}+\bm{\delta}) \approx \mathcal{L}(\bm{x}) + \sum_i^d \nabla_{x_i}\mathcal{L}(\bm{x})\delta_i + \frac{1}{2}\sum_{i, j}^d \nabla_{x_ix_j}^2\mathcal{L}(\bm{x})\delta_i\delta_j
    \label{equ: adversarial loss}
\end{equation}
where $\nabla_{x_i}\mathcal{L}(\bm{x})$, or $\nabla_{x_i}$ for short, is the first-order gradient of $\mathcal{L}(\bm{x})$ w.r.t. the input dimension $x_i$ corresponding to the slope of the input loss landscape. Similarly, $\nabla_{x_ix_j}^2\mathcal{L}(\bm{x})$, or $\nabla_{x_ix_j}^2$ for short, denotes the second-order gradient w.r.t. $x_i$ and $x_j$ corresponding to the curvature of the loss landscape. 

We define the adversarial vulnerability (AV) of $\bm{x}$ against a particular attack as the loss increment caused by the attack:
\begin{equation}
    \text{AV} = \mathcal{L}(\bm{x}+\bm{\delta}) - \mathcal{L}(\bm{x})
\label{equ: adversarial vulnerability}
\end{equation}
A higher vulnerability means that adversarial attack has a greater impact on the loss value for that sample, and hence, is more likely to corrupt the model's prediction for that sample.
From this perspective, loss gradients in \cref{equ: adversarial loss} constitute the source of AV. Adversarial attacks exploit input loss gradients to enlarge adversarial loss by aligning the sign of the perturbation and corresponding gradient, i.e., $sign(\delta_i)=sign(x_i)$. Gradients with small magnitude (a flat loss landscape) therefore indicate low AV. 

AT works by shrinking the magnitude of gradients, or flattening the loss landscape from the geometric perspective. Taking a toy example of training with two samples, adversarial loss can be written (second-order gradients are ignored for simplicity) as
\begin{equation}
    \mathcal{L}(\bm{x}+\bm{\delta}) \approx \mathcal{L}(\bm{x}) + \sum_i^d \nabla_{x_{1,i}}\delta_{1,i} + \sum_i^d \nabla_{x_{2,i}}\delta_{2,i}
    \label{equ: two samples adversarial loss}
\end{equation}
Supposing that the training adversary is theoretically optimal, i.e., $\forall i, j: sign(\nabla_{x_{i,j}}) = sign(\delta_{i,j})$ and $\left|\delta_{i,j}\right| = \epsilon$, the objective of AT can be rewritten as
\begin{equation}
    \arg \min_{\bm{\theta}}\ \mathcal{L}(\bm{x}) + \epsilon\sum_i^d \left|\nabla_{x_{1,i}}\right| + \epsilon\sum_i^d \left|\nabla_{x_{2,i}}\right|
    \label{equ: two samples adversarial training objective}
\end{equation}
Note $\left|\ \cdot\ \right|$ denotes the absolute value. Ideally, the magnitude of every gradient is supposed to be shrunk by AT with the decrease of training loss resulting in a flat loss landscape over the input space for all training data. 

However, \cref{equ: two samples adversarial training objective} can also be reduced  by sacrificing some gradients (to be large) in order to shrink other gradients or the clean loss: as long as the overall reduction is greater than any enlargement. From the geometric view, the loss landscape becomes steep around some samples so that it can be flat and/or low around others. If some gradients are consistently sacrificed (enlarged), the model may eventually converge to yield an uneven distribution of AV among training data.

We argue that it is even more likely for the model to converge to such an uneven state if the training adversary is weaker than the theoretical optimum. A sub-optimal adversary causes misalignment between the sign of the perturbation and the corresponding gradient, such that $sign(\nabla_{x_{i,j}}) \neq sign(\delta_{i,j})$ in \cref{equ: two samples adversarial loss}, in which case training encourages the misaligned gradients to grow larger. If some gradients are consistently misaligned by their corresponding perturbations, they accumulate to be large.

\begin{figure}[tbp]
\subfloat[]{
    \includegraphics[width=0.9\columnwidth]{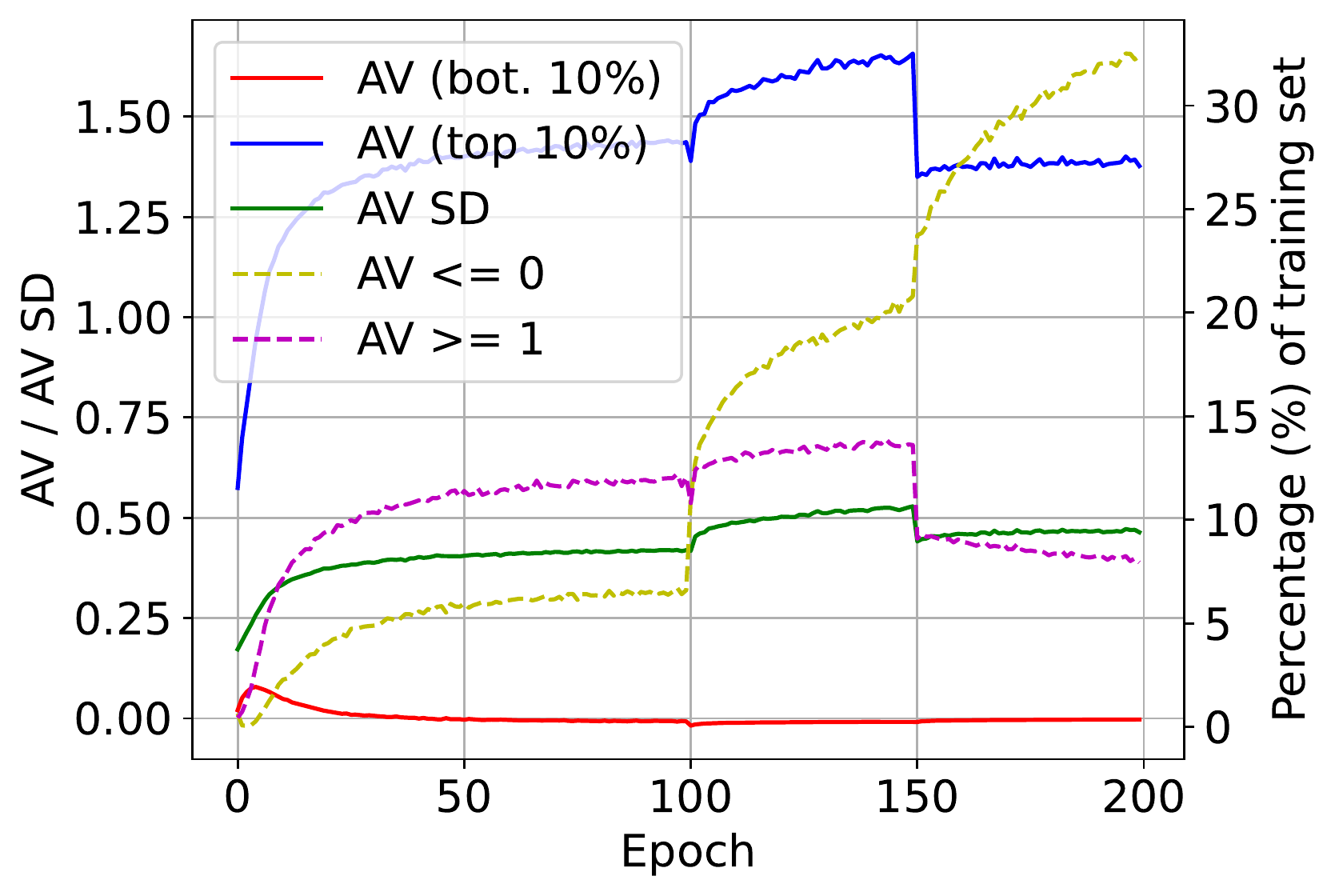}
    \label{fig: av statistics over adversarial training}
}

\subfloat[]{
    \includegraphics[width=0.9\columnwidth]{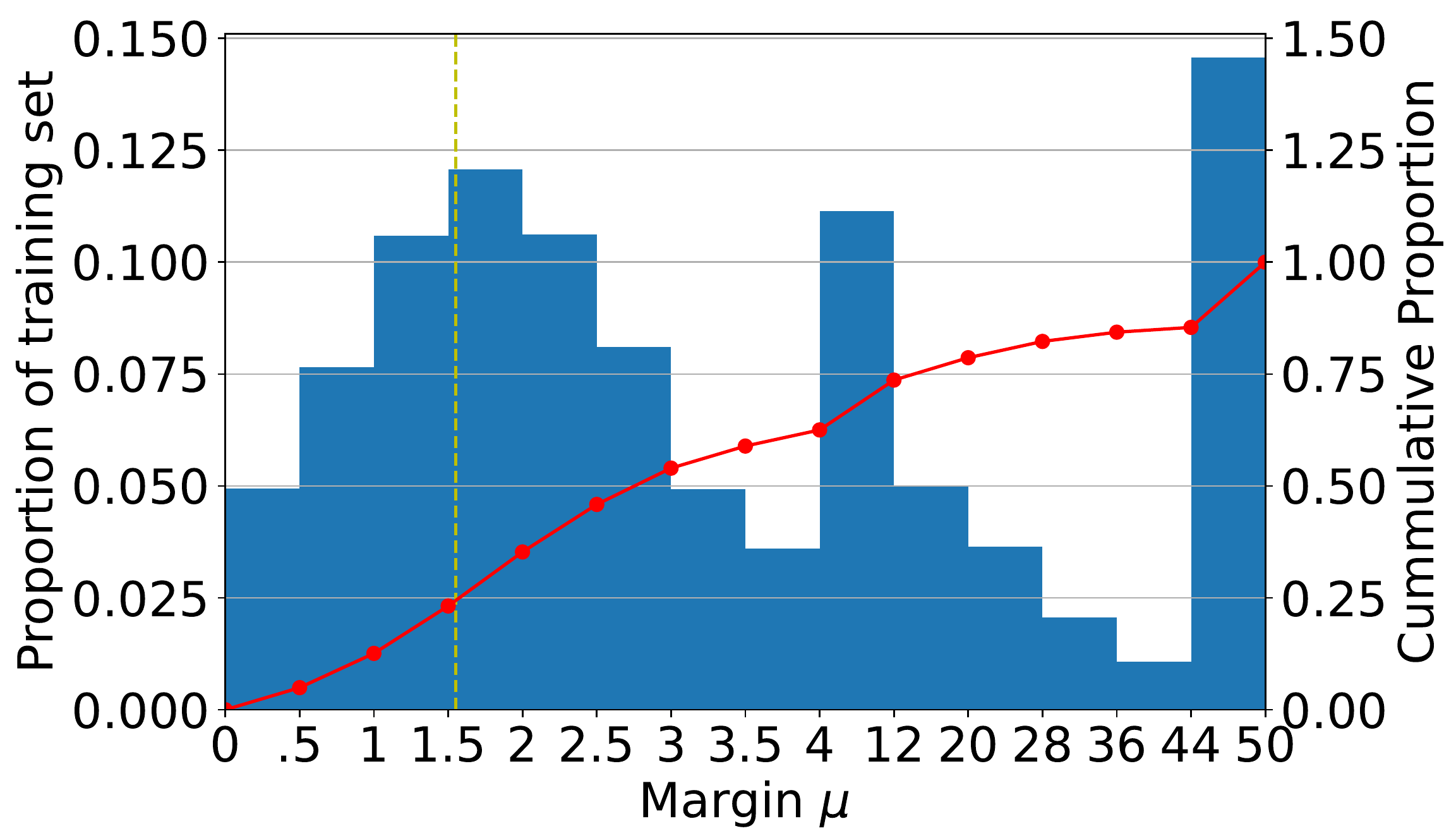}
    \label{fig: margin distribution}
}
\caption{Illustration of the phenomenon "uneven vulnerability".
\textbf{\cref{fig: av statistics over adversarial training}} shows how instance-wise AV evolves with the training epoch. "AV (bot. 10\%)" and "AV (top 10\%)" measure the average AV of the 10\% of training samples with the lowest and highest AV respectively. "AV SD" is the standard deviation of AV across all training samples. "AV $\leq$ 0" and "AV $\geq$ 1" measure the proportion of samples with AV $\leq 0$ and $\geq 1$ among the entire training set.
\textbf{\cref{fig: margin distribution}} shows the distribution of margin $\mu$ over the entire training set for the checkpoint with highest robustness found in the experiment of \cref{fig: av statistics over adversarial training}. The yellow dashed line indicates the margin corresponding to the training perturbation budget $\epsilon$. The red solid line represents cumulative distribution.}
\end{figure}

% \begin{figure}
% \captionsetup[subfigure]{labelformat=empty}
%     \hfill
%     \subfloat[Airplane: .965]{
%         \includegraphics[width=.22\linewidth]{figure/excessive-margin-img-vis/0_cle_0.965.png}
%     }
%     \subfloat[Airplane: .976]{
%         \includegraphics[width=.22\linewidth]{figure/excessive-margin-img-vis/0_adv_0.976.png}
%     }
%     \subfloat[Frog: .994]{
%         \includegraphics[width=.22\linewidth]{figure/excessive-margin-img-vis/6_cle_0.994.png}
%     }
%     \subfloat[Frog: .942]{
%         \includegraphics[width=.22\linewidth]{figure/excessive-margin-img-vis/6_adv_0.942.png}
%     }
%     \hfill
%     \label{fig: visualization of ill robust samples}
%     \caption{Visualization of disorded robust samples and their adversarial examples corresponding to margin $\mu=30$ (about 20 times of $\epsilon$). The caption of each image describes the ground-truth class and the prediction confidence (the output of softmax layer at the index of ground-truth class) in order. Although the images are modified dramatically by adversarial perturbation to a detrimental degree, they can still be correctly classified by the model without much variation on confidence.}
% \end{figure}

\begin{figure}[tbp]
    \centering
    \subfloat{
        \includegraphics[width=\linewidth, trim=15 30 0 0]{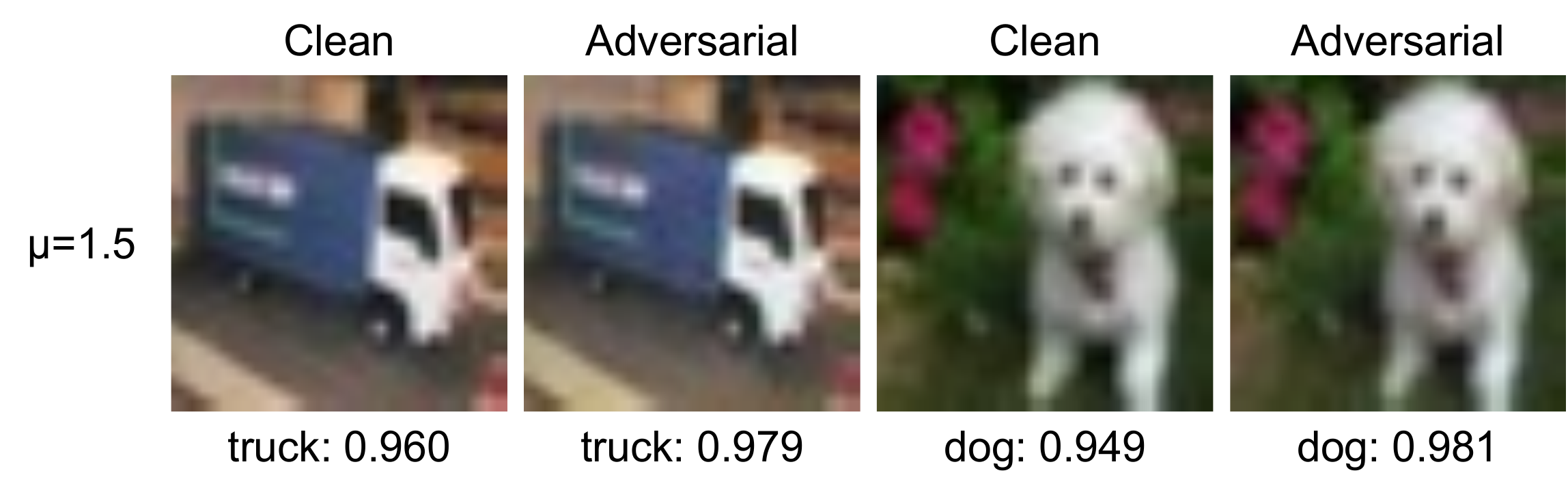}
    }

    \subfloat{
        \includegraphics[width=\linewidth, trim=15 30 0 0]{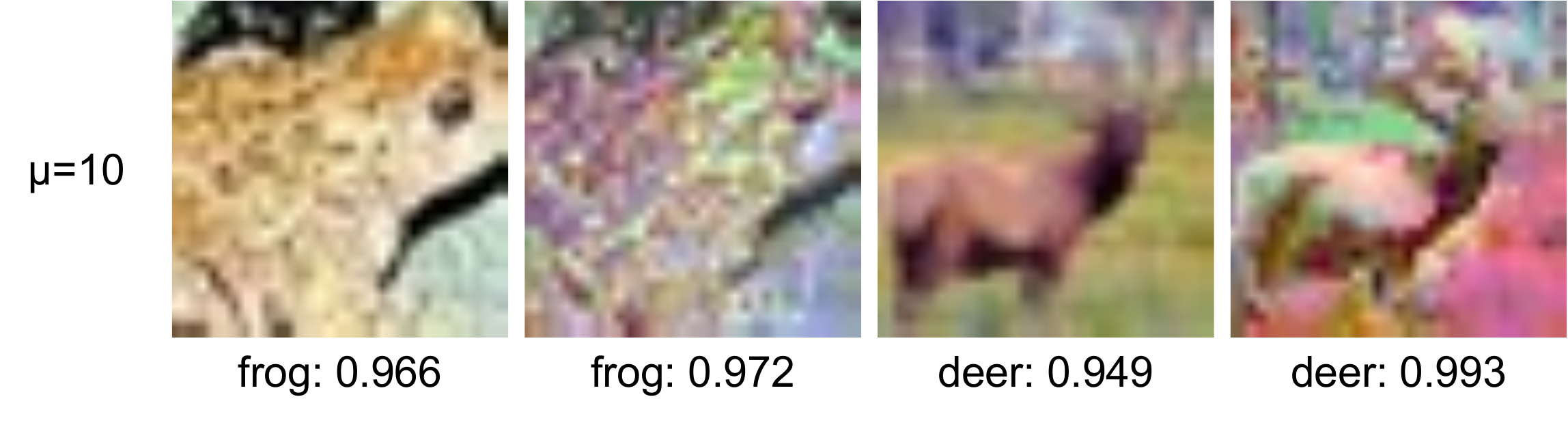}
    }

    \subfloat{
        \includegraphics[width=\linewidth, trim=15 30 0 0]{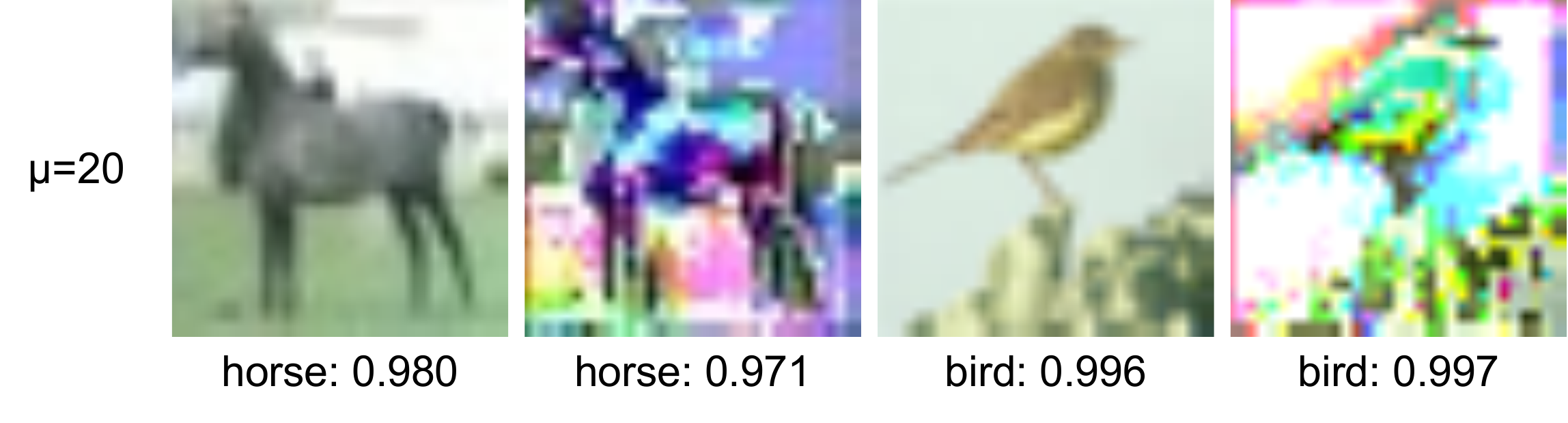}
    }

    \subfloat{
        \includegraphics[width=\linewidth, trim=15 30 0 0]{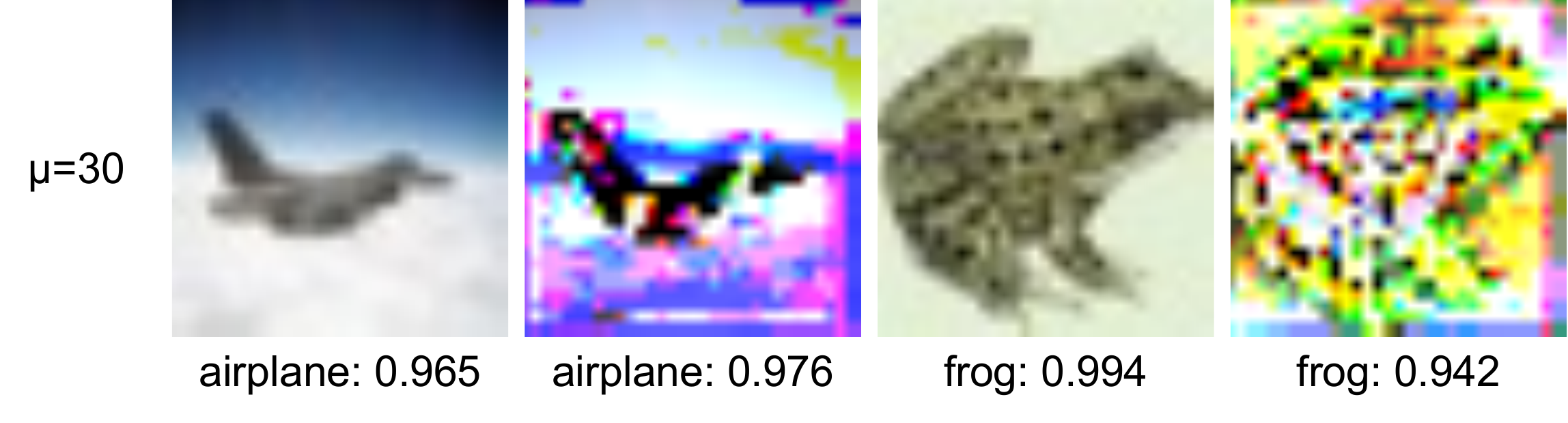}
    }

    \subfloat{
        \includegraphics[width=\linewidth, trim=15 30 0 0]{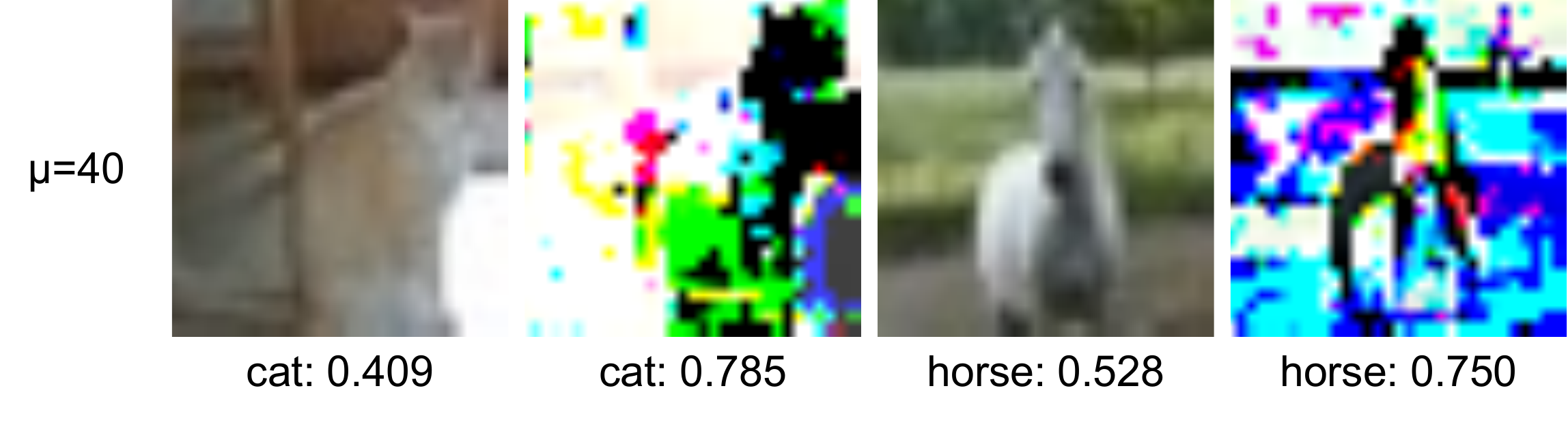}
    }

    \subfloat{
        \includegraphics[width=\linewidth, trim=15 20 0 0]{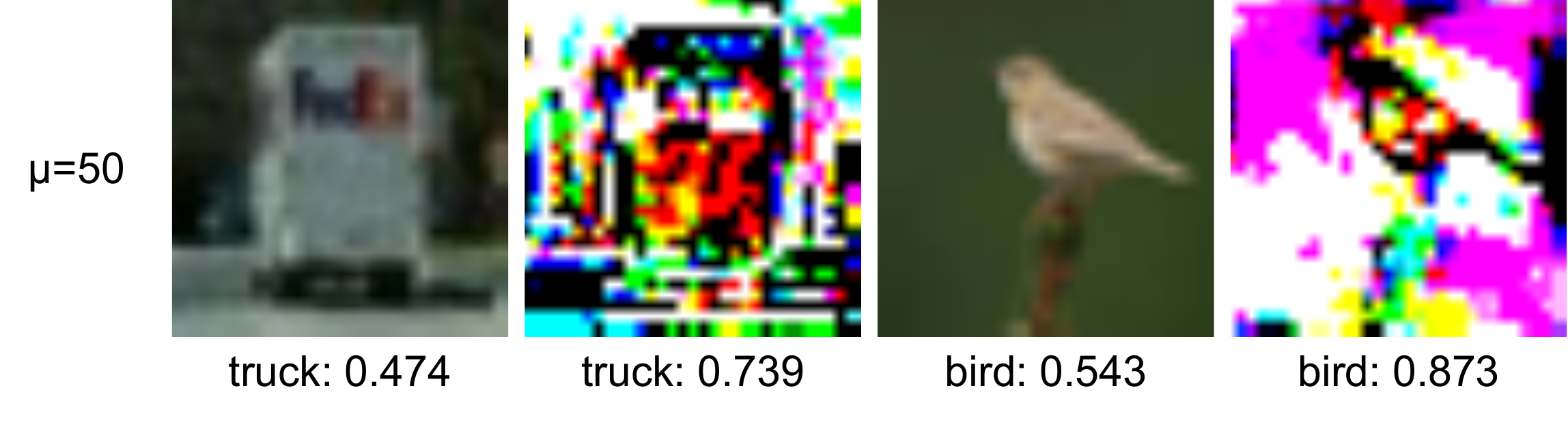}
    }    
    \caption{Visualization of disordered robustness. This figure shows clean samples and the corresponding adversarial samples for various levels of margin. Adversarial examples in each row were crafted using $\bm{x}+\mu \frac{\bm{\delta}}{\lVert \bm{\delta} \rVert_2}$ with the value of $\mu$ given at the left of each row. $\mu=1.5$ approximately corresponds to the perturbation budget $\epsilon$. The caption of each image describes the ground-truth class and the prediction confidence (the output of softmax layer at the index of ground-truth class). Although the images are modified dramatically by adversarial perturbation to even a detrimental degree when $\mu \geq 20$, they can still be correctly classified by the model without much variation on confidence, or even with higher confidence. In the cases with very large $\mu$, the perturbed images become extremely hard to recognize or become meaningless to a human observer, while the model is able to recognize them correctly with high confidence.}
    \label{fig:visualization_of_ill_robust_samples}
\end{figure}

\begin{figure*}[tbp]
\subfloat[]{
\includegraphics[width=.23\linewidth, trim=10 0 20 20]{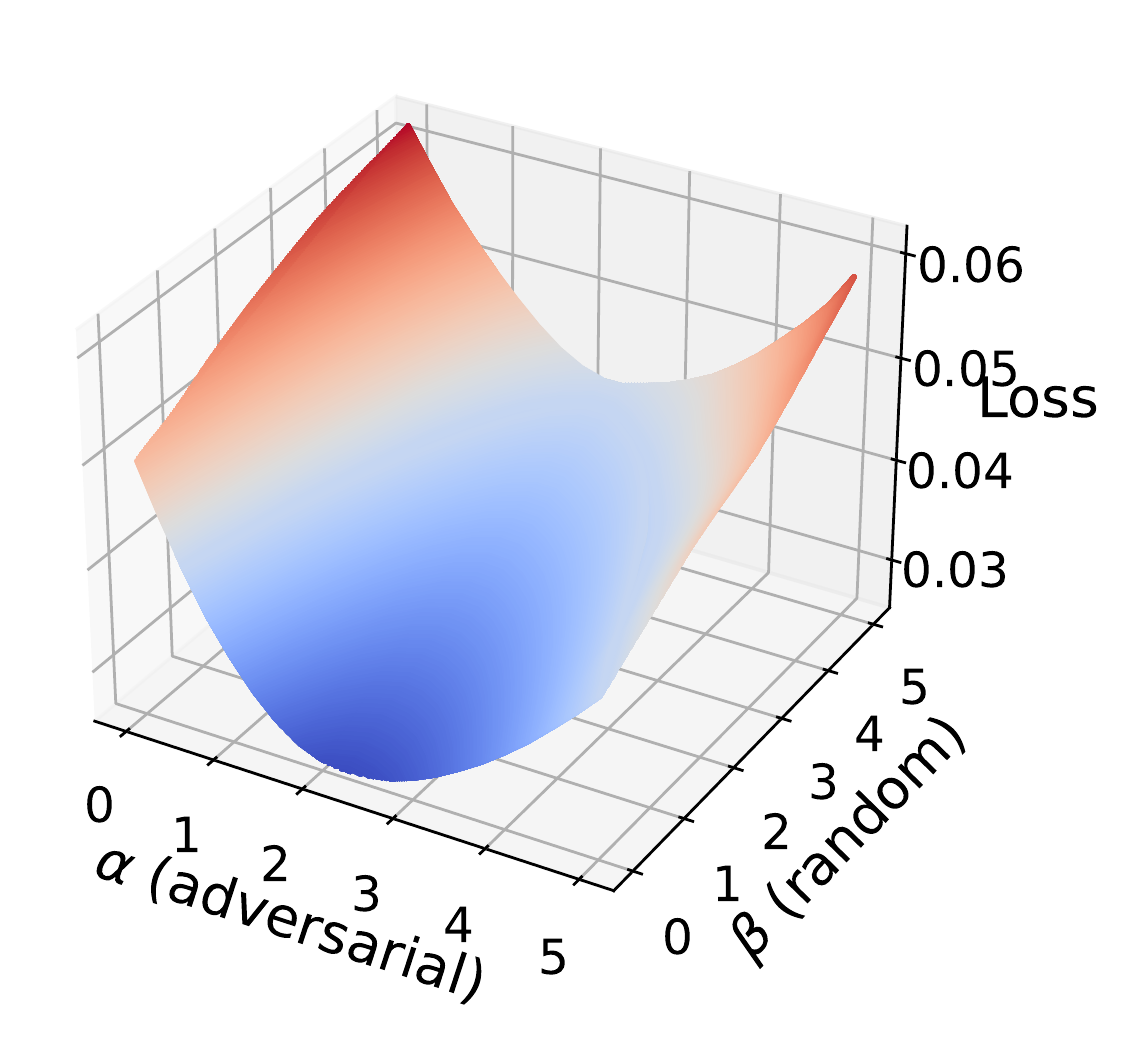}
\label{fig: disordered robustness loss vis 1}
}
\hfill
\subfloat[]{
\includegraphics[width=.23\linewidth, trim=10 0 20 20]{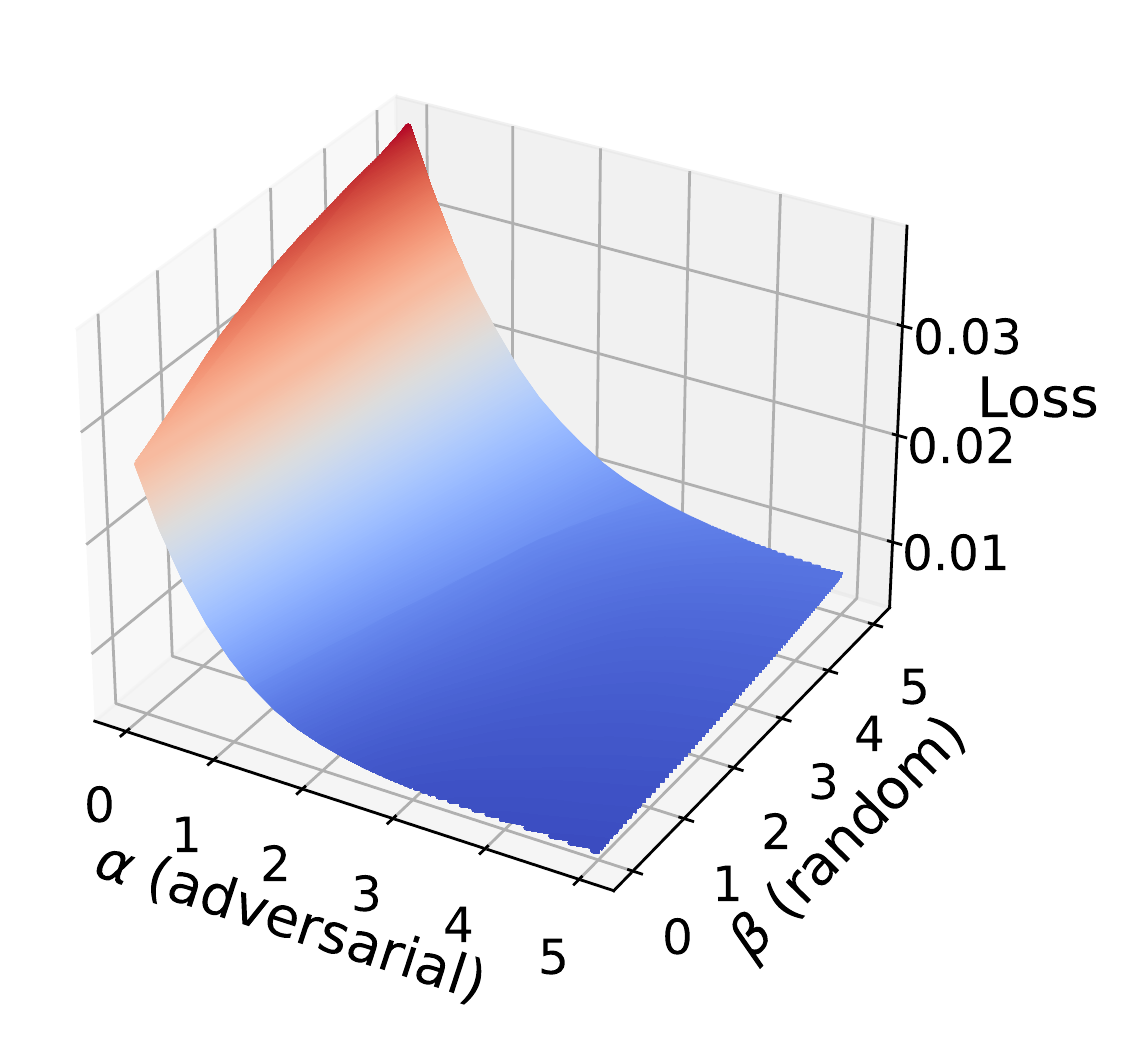}
\label{fig: disordered robustness loss vis 2}
}
\hfill
\subfloat[]{
\includegraphics[width=.23\linewidth, trim=10 0 20 20]{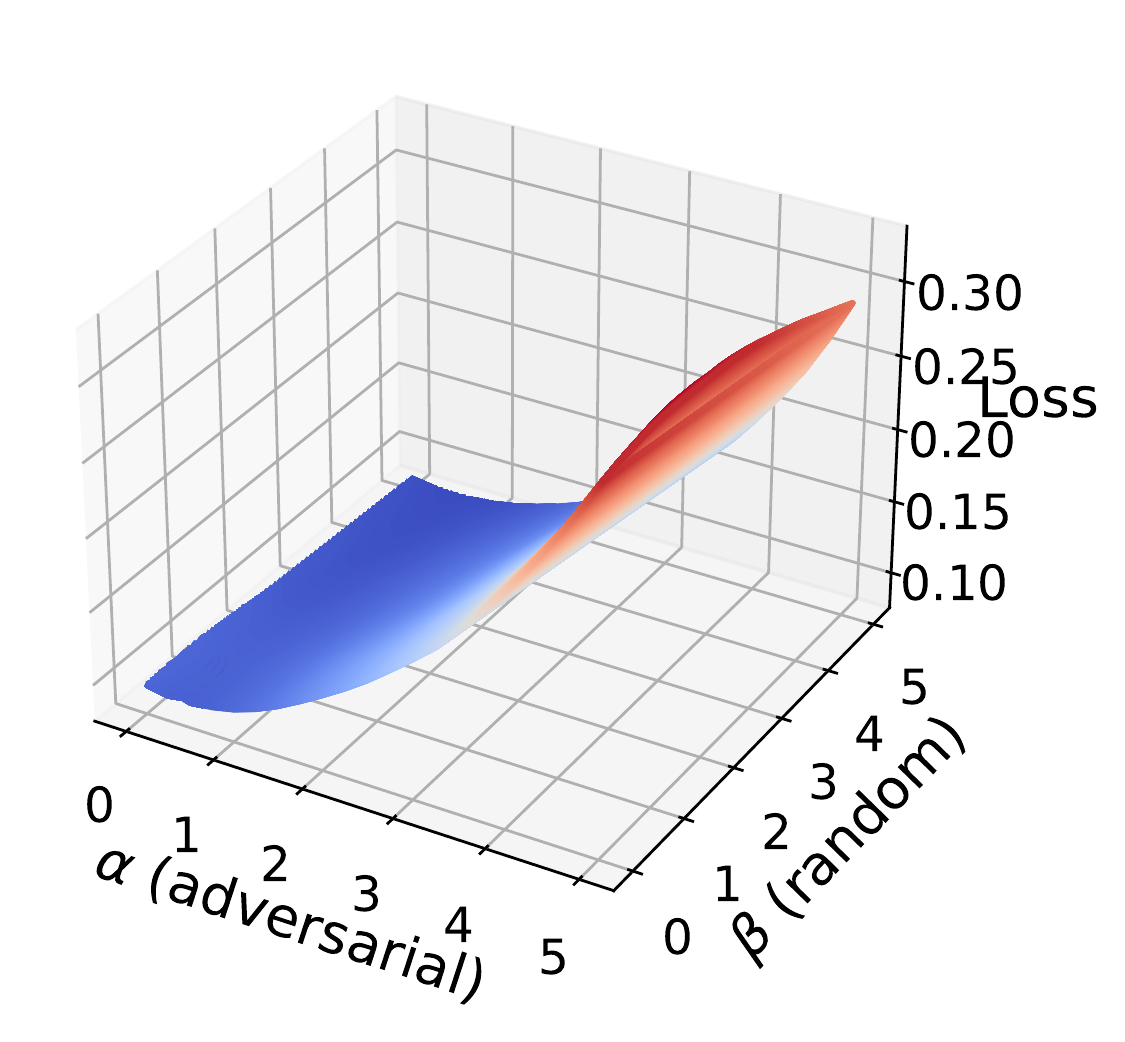}
\label{fig: robust loss vis}
}
\hfill
\subfloat[]{
\includegraphics[width=.23\linewidth, trim=10 0 20 20]{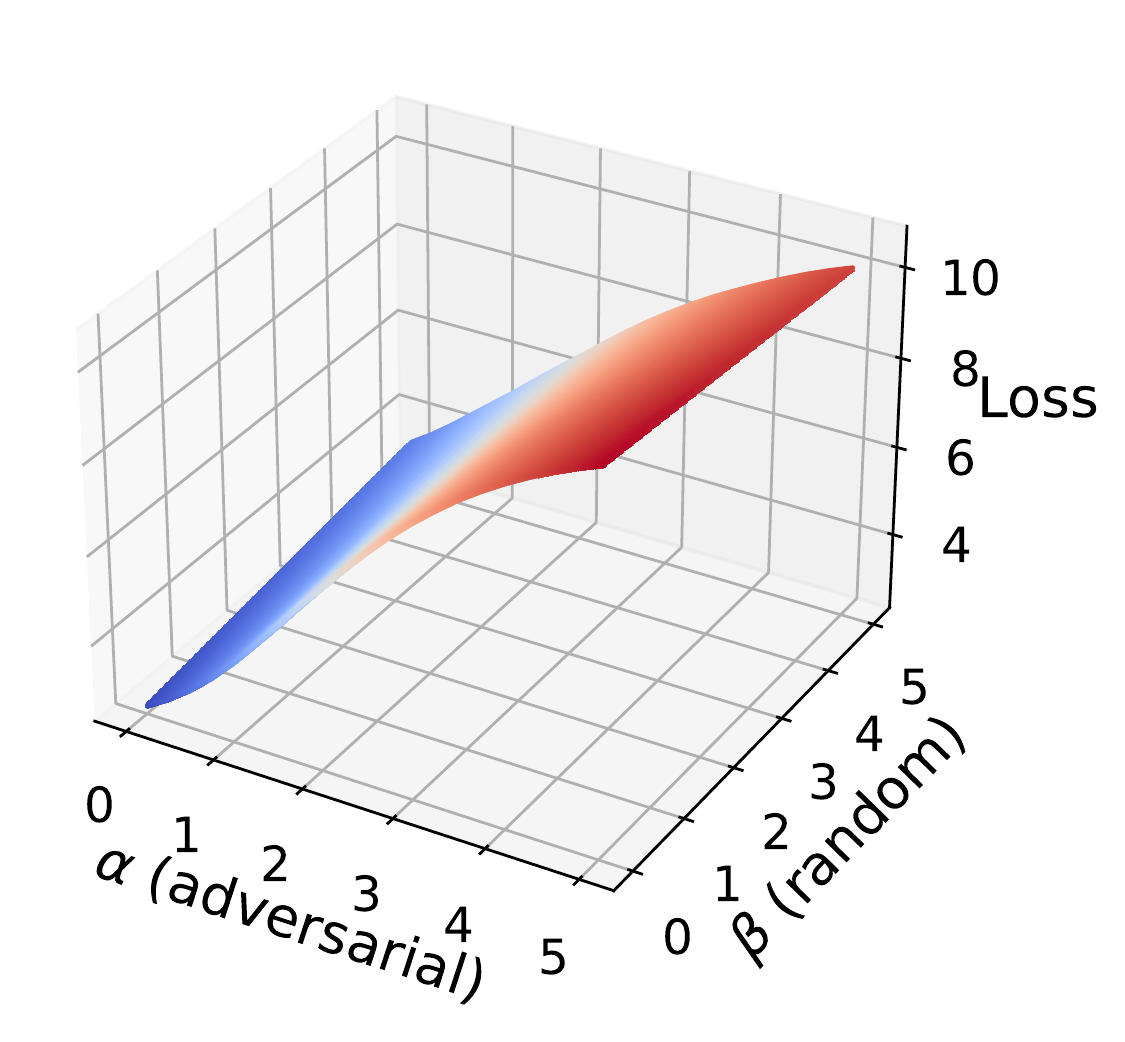}
\label{fig: vulnerable loss loss vis}
}

\caption{Visualization of loss landscapes at samples with disordered robustness (\cref{fig: disordered robustness loss vis 1} and \cref{fig: disordered robustness loss vis 2}), and for a robust sample (\cref{fig: robust loss vis}) and a vulnerable sample (\cref{fig: vulnerable loss loss vis}). 
$\alpha \approx 1.5$ corresponds to a perturbation size of $\epsilon$.
Loss increases up as the color of plane changes from blue to red. Note the scale of loss is dramatically different for these three categories of data.}
\label{fig: loss vis}
\end{figure*}

\subsection{Uneven Distribution of Adversarial Vulnerability} \label{sec: uneven distribution of adversarial vulnerability}

AT in practice is prone to minimize training loss in the alternative way described in the final paragraphs of the previous section. Consequently, AV is unevenly distributed among the data. To demonstrate this we track how the instance-wise AV of the training data varies as training progresses. Specifically, we measure the degree of unevenness for AV via its standard deviation (SD) over the entire training set:
\begin{equation}
    \text{AV SD} = \sqrt{\mathbb{E}_{i \in D_t}[(\text{AV}_i-\mathbb{E}_{j \in D_t}[\text{AV}_j])^2]}
\end{equation}
Higher AV SD indicates that AV is spread out more among instances, i.e., higher unevenness. In addition, we compute the mean AV of the 10\% of samples with the lowest AV ("bottom 10\%") and the mean AV of the 10\% of samples with the highest AV ("top 10\%"). We also assess the unevenness of adversarial vulnerability by calculating the percentage of samples with AV $\geq 1$ and $\leq 0$ within
the whole training set. $1$ and $0$ are selected as the thresholds for high and low AV respectively as the model's prediction, after adversarial attack, is observed to be significantly affected if AV $\geq 1$ and hardly changed if AV $\leq 0$.

%AV disperses among instances over AT. 
In \cref{fig: av statistics over adversarial training} it can be observed that AV SD, as well as the number of high- and low-vulnerability samples, increases over the course of training. The average AV of top 10\% increases, while the average AV of bottom 10\% decreases to be even lower than 0. Although the overall training adversarial loss decreases, AV is distributed unevenly among training samples: some become extremely vulnerable to attack while others become very robust.

To further examine this phenomenon, the margin ($\mu$) along the adversarial direction 
for each sample in the training data was computed from the adversarially-trained model
using the method defined in \cite{rade_reducing_2022}:
\begin{equation}
    \arg \min_{\mu}\ \text{s.t.}\ F(\bm{x}+\mu \frac{\bm{\delta}}{\lVert \bm{\delta} \rVert_2}; \bm{\theta}) \neq F(\bm{x} ; \bm{\theta})
\end{equation}
Where $\bm{\delta}$ is computed using PGD50 and $\lVert \cdot \rVert_2$ is the $\ell_2$-norm.
As shown in \cref{fig: margin distribution}, about 20\% of training data can be successfully attacked within the $\epsilon$-ball to fool the model into changing prediction, and among them, around 5\% of training instances can be successfully attacked using only a third of the perturbation budget $\epsilon$. In contrast, a large proportion  of samples exhibit an excessive margin along the adversarial direction. The prediction of the model remains constant under an attack with double the perturbation budget for about half the training data. More surprisingly, about 14\% of the training samples exhibit the theoretically maximal effective margin ($\mu = 50$)\footnote{The margin value corresponding to the perturbation budget $\epsilon$ is about 1.5. A margin value of 50 is, hence, equivalent to perturbing along the adversarial direction by approximately $\frac{50}{1.5}\epsilon$ which is greater than 1. For our model, input images are normalized to have pixel values between zero and one, and the perturbed input is clipped to remain in this range, so increasing the magnitude of any perturbation beyond a value of one with have no additional effects on the input.}, which indicates that no perturbation along the adversarial direction can change the model's prediction. 
We name this property of extremely excessive margin as "disordered robustness" because a reasonable model should be sensitive to noticeable or even devastating perturbations of the input (see \cref{fig:visualization_of_ill_robust_samples} for examples of perturbed inputs for sample with disordered robustness). 

To further verify our claim, the loss landscapes around some representative samples are visualized in \cref{fig: loss vis}. The loss was calculated using:
\begin{equation}
    L = \mathcal{L}(\bm{x}+\alpha\frac{\bm{\delta}}{\lVert \bm{\delta} \rVert_2} + \beta \frac{\bm{u}}{\lVert \bm{u} \rVert_2}, y; \bm{\theta})
\label{equ: loss landscape visualization}
\end{equation}
Where $\bm{\delta}$ was generated by PGD50 and $\bm{u}$ was randomly sampled from a uniform distribution $\mathcal{U}(-\epsilon, \epsilon)^d$. The loss landscape was visualized along the adversarial and the random direction by varying $\alpha$ and $\beta$ respectively.

For samples with disordered robustness, lower loss values are produced for values of $\alpha>0$ than are produced when $\alpha=0$ (see \cref{fig: disordered robustness loss vis 1} and \cref{fig: disordered robustness loss vis 2}). This confirms that this particular kind of robustness is disordered because adversarial examples are more benign, i.e., easier to be correctly classified than clean examples in this case. In fact, perturbations in the random direction are more malicious than those in the adversarial direction since the loss increases as $\beta$ increases. Nevertheless, the highest loss value achieved by perturbations along the random direction is still very small, i.e., the perturbed sample remains easy to be correctly classified. Therefore, such disordered robust samples are very resistant to the input perturbations within the budget. 

In contrast, the loss landscapes for (normal) robust samples (\cref{fig: robust loss vis}) and vulnerable samples (\cref{fig: vulnerable loss loss vis}) are quite different. It can be observed that both robust and vulnerable samples exhibit an increasing loss as $\alpha$ is increased (i.e. as the magnitude of the perturbation along the adversarial direction is increased), which is in stark contrast to the loss landscape at samples with disordered robustness.

We acknowledge that excessive margin has been observed before in AT \cite{rade_reducing_2022}. However, our finding differs regarding the direction along which excessive margin is observed. \cite{rade_reducing_2022} observed excessive margin for an adversarially-trained model along the adversarial direction found by PGD on a standardly-trained model (i.e. they used different models for the adversarial direction and the margin evaluation), while we observed it along the PGD adversarial direction generated for an adversarially-trained model (same model for adversarial direction and margin evaluation). 
Furthermore, \cite{rade_reducing_2022} did not find that the direction along which the excessive margin is observed is in fact adversarially benign. 
% Furthermore, the phenomenon of "benign" adversarial direction is novel and has been never observed before in a normal, robust, model.

% TODO: prevalence in other datasets and models in Appendix

\subsection{Connection of Uneven Adversarial Vulnerability to Robust Overfitting} \label{sec: connection to robust overfitting}
We hypothesize that the uneven distribution of AV accounts for overfitting in AT. 
% Intuitively, higher unevenness implies that some samples are relatively more vulnerable
To evaluate, we test how robust generalization varies with the unevenness of AV. Unevenness is controlled by the strength, $\eta$, of a logit stability regularization applied to the 10\% of samples with the highest AV:
\begin{equation}
    \mathbb{E}_{i \in M}[\mathcal{L}(\bm{x}_i+\bm{\delta}_i) + \eta\lVert f(\bm{x}_i+\bm{\delta}_i)-f(\bm{x}_i) \rVert_2^2 \mathbbm{1}(\bm{x}_i, \bm{\delta}_i)]
\label{equ: topn regularization}
\end{equation}
where $\mathbbm{1}(\cdot)$ is an indicator function to select the samples with highest AV within each training batch. $\lVert \cdot \rVert_2^2$ is the squared $\ell_2$-norm. Unevenness is supposed to be reduced as $\eta$ increases since those highly vulnerable samples are regularized to be more robust. We use this regularizer to fine-tune a pre-adversarially-trained model for 10 epochs with an initial learning rate of 0.01 decayed by 0.1 at half epochs. Experiments are preformed using different values for $\eta$ from 0 to 4 with a step size of 0.5. Note that $\eta=0$ means no regularization is applied.

\begin{figure}[tbp]
    \centering
    \includegraphics[width=\linewidth,trim=8 8 7 0, clip]{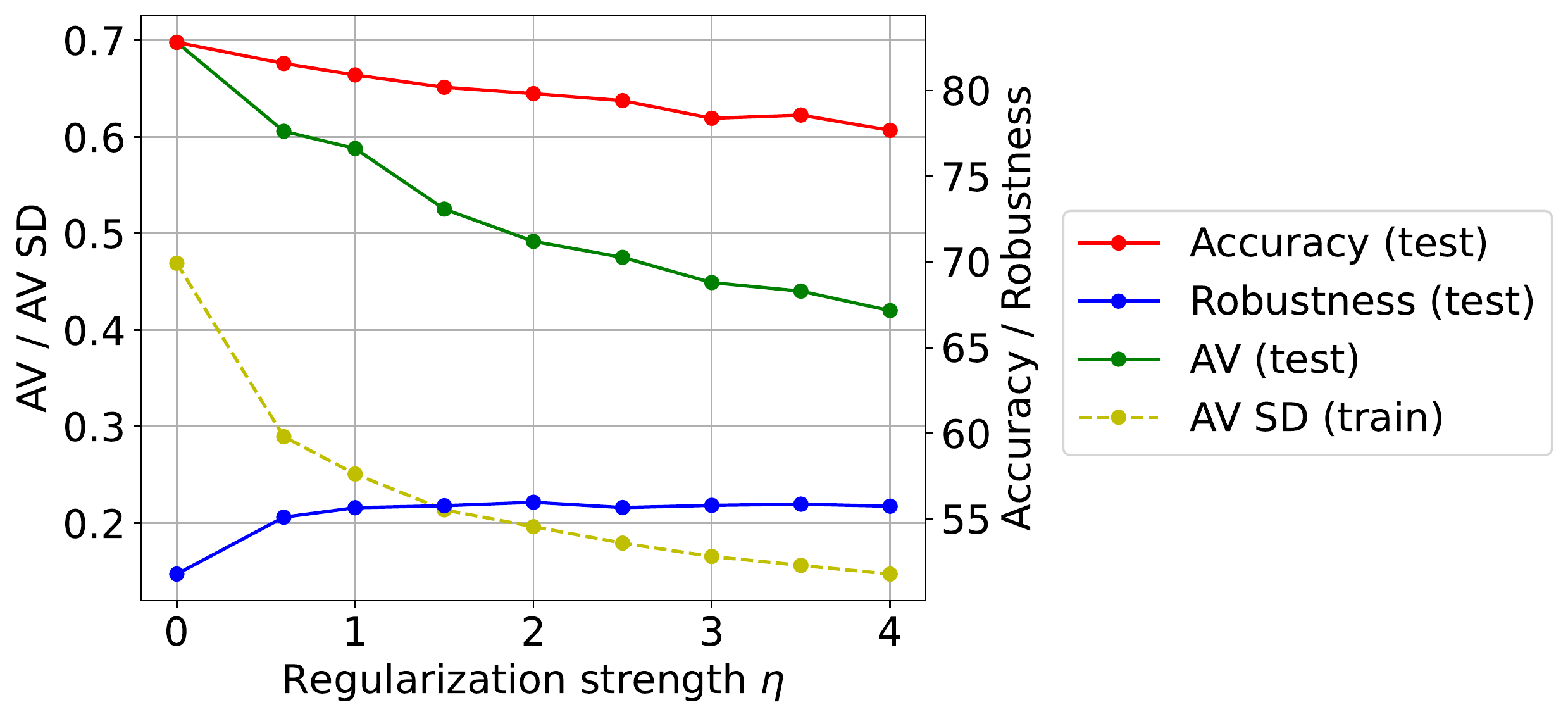}
    \caption{Accuracy, robustness and  adversarial vulnerability for the models trained with different regularization strength. For training data, we measure AV SD to indicate how the unevenness of adversarial vulnerability varies. Regarding test performance, we evaluate accuracy, robustness and adversarial vulnerability (AV) on the test set.
    % Accuracy and robustness 
    % are evaluated on the test set, while AV SD is evaluated for both the test and training data. %are evaluated on the test set, while unevenness is evaluated on the training set.
    }
    \label{fig: connection to ro}
\end{figure}

As shown in \cref{fig: connection to ro}, the AV SD on the training data drops with increasing regularization strength. 
The AV of the test data decreases even more dramatically, and test robustness increases, indicating improved robust generalization with stronger regularization. The increase of test robustness is less significant than the decrease of test AV because clean accuracy degrades with stronger regularization. Overall, this result verifies that the unevenness of AV is related to the degree of robust overfitting. Note that this does not imply that uneven AV is necessary for robust overfitting. For example, it is possible for a model with extremely large capacity to perfectly overfit every training data in AT so that all training data have nearly $0$ AV. Nevertheless, uneven vulnerability might still occur at an intermediate stage during learning, and reduce as learning continues to produce perfect robust overfitting. Similarly, we observe that unevenness declines after the second decay of learning rate in \cref{fig: av statistics over adversarial training}.

\subsection{Prevalence of Uneven Adversarial Vulnerability Across Robust Training Schemes} \label{sec: prevalence across various robust training schemes}
Finally, we show that the issue of unevenly distributed AV is prevalent across various robust training schemes. As shown in \cref{tab: av across adversarial training methods}, RST and AWP both mitigate the unevenness of AV to some extent with a reduced top 10\% average AV and reduced number of high-vulnerability samples compared to vanilla AT. However, the reduction of unevenness produced by RST and AWP is less than that achieved by the purpose-built regularization (\cref{equ: topn regularization}). This suggests that these previous methods of improving AT contribute to enhanced adversarial robustness using different mechanism to the proposed one and a higher robustness can be expected if they are properly combined together, as described next.

% it seems an issue of the data itself: among all evaluated settings, some samples are always sacrificed, while some are always over flat.

% \usepackage{booktabs}
% \usepackage{multirow}
% \usepackage{graphicx}
% \usepackage[normalem]{ulem}
\begin{table}[tbp]
\centering
\caption{Robustness and the statistics of adversarial vulnerability for various robust training schemes. "AV SD" denotes the standard deviation of adversarial vulnerability. "Top 10\%" ("Bot. 10\%") denotes the 
the average AV of the 10\% of training samples with the highest and lowest AV respectively. "$\geq 1$" ("$\leq 0$") denotes the proportion of training data with adversarial vulnerability greater (less) than or equal to 1 (0).}
\label{tab: av across adversarial training methods}
%\resizebox{\columnwidth}{!}{%
\begin{tabular}{@{}lcccccc@{}}
\toprule
\multirow{2}{*}{Method} &
  \multirow{2}{*}{Rob. (\%)} &
  \multicolumn{5}{c}{Adversarial vulnerability} \\ \cmidrule(l){3-7} 
\multicolumn{1}{c}{} &                      & AV SD    & Top 10\% & Bot. 10\% & $\geq 1$     & $\leq 0$     \\ \midrule
AT                   & 51.78                & 0.467 & 1.527    & -0.010    & 12.38 & 12.14 \\
AWP                  & 54.68                & 0.351 & 1.120    & -0.022    & 5.52  & 19.74 \\
RST                  & {\ul \textbf{57.68}} & 0.443 & 1.378    & -0.011    & 9.96  & 20.87 \\
\cref{equ: topn regularization} &
  55.95 &
  {\ul \textbf{0.196}} &
  {\ul \textbf{0.633}} &
  {\ul \textbf{-0.047}} &
  {\ul \textbf{0.14}} &
  {\ul \textbf{12.00}} \\ \bottomrule
\end{tabular}%
%}
\end{table}\tabcolsep\tabcolseporig

\section{Instance Adaptive Smoothness Enhanced Adversarial Training}

Inspired by the above insights, we propose a new general AT paradigm, Instance Adaptive Robustness Enhancement, that alleviates the uneven vulnerability issue to improve robust generalization. This approach enhances robustness for training samples with a strength adaptive to their vulnerability. In general, a higher strength of regularization is applied to samples with higher vulnerability with the aim of improving robustness for these samples. In contrast, low-vulnerability samples, which are already robustly classified, receive weaker regularization due to the desire to avoid the effects of over-regularization. Specifically, we combine AT with a robustness regularizer and adapt the regularization strength for each instance based on its AV. \cref{alg: general framework} illustrates this general framework.
%To realize a specific solution, we first design a linear weight scheme based on the relative order of sample's AV. 

%Next, we propose a new regularizer as the robustifying mean in the paradigm that jointly smooths input and weight loss landscapes. Overall, we have our ultimate solution of Instance adaptive Smoothness Enhanced Adversarial Training (ISEAT).

%\subsection{Instance Adaptive Robustness Enhancement}

Our ultimate proposal for improving AT is a specific implementation of this general framework that uses AV weighted regularization to jointly smooth the input and weight loss landscapes. This specific realization of our proposal is called Instance-adaptive Smoothness Enhanced Adversarial Training (ISEAT), and is summarized in \cref{alg: instance-wise smoothness enhanced adversarial training} and described in detail below.

First, in order to integrate weight loss smoothing into our framework, as described in detail in the following section, we extend the metric defined in \cref{equ: adversarial vulnerability} to measure instance-wise vulnerability against adversarial input, $\bm{\delta}$, and weight, $\bm{v}$, perturbation:
\begin{equation}
    \text{AV}_i = \mathcal{L}(\bm{x}_i+\bm{\delta}_i; \bm{\theta}+\bm{v}) - \mathcal{L}(\bm{x}_i; \bm{\theta})
\label{equ: adversarial vulnerability against input and weight perturbation}
\end{equation}
$\bm{v}$ is an adversarial perturbation within a pre-defined feasible region, $\mathcal{V}$, applied to the model's parameters to maximize adversarial loss (adversarial input perturbation $\bm{\delta}$ is assumed) \cite{wu_adversarial_2020}:
\begin{equation}
    \arg \max_{\bm{v} \in \mathcal{V}}\ \mathbb{E}_{i \in M} [\mathcal{L}(\bm{x}_i+\bm{\delta}_i; \bm{\theta}+\bm{v})] 
\label{equ: v objective}
\end{equation}
To optimize, $\bm{v}$ is searched like $\bm{\delta}$ by the projected gradient descent (PGD) algorithm \cite{madry_towards_2018} as: 
\begin{equation}
    \bm{v} \leftarrow \Pi_{\gamma}(\bm{v} + \rho \frac{\nabla_{\bm{v}} \mathbb{E}_{i \in M}[\mathcal{L}(\bm{x}_i+\bm{\delta}_i;\bm{\theta}+\bm{v})]}{\lVert \nabla_{\bm{v}} \mathbb{E}_{i \in M}[\mathcal{L}(\bm{x}_i+\bm{\delta}_i;\bm{\theta}+\bm{v})] \rVert} \lVert \bm{\theta} \rVert )
\end{equation}
where $\rho$ is the step size, and $\Pi_{\gamma}(\cdot)$ is a layer-wise projection operation, defined as follows:
\begin{equation}
    \Pi_{\gamma}(\bm{v}) = \begin{cases}
        \gamma \frac{\lVert \bm{\theta}_n \rVert}{\lVert \bm{v}_n \rVert}\bm{v}_n, &\text{if} \lVert \bm{v}_n \rVert > \lVert \gamma \bm{\theta}_n \rVert, \forall n \in N \\
        \bm{v}_n, &\text{if} \lVert \bm{v}_n \rVert \leq \lVert \gamma \bm{\theta}_n \rVert, \forall n \in N
    \end{cases}
\label{equ: projection gamma}
\end{equation}
where $N$ denotes the collection of all layers in the network. 
$\Pi_{\gamma}(\cdot)$ restricts the weight perturbation $\bm{v}_n$ in the $n$-th layer to be relative to the corresponding weight $\bm{\theta}_n$ such that $\lVert \bm{v}_n \rVert \leq \lVert \gamma \bm{\theta}_n \rVert$. For more detail of this relative perturbation size, please refer to the original work \cite{wu_adversarial_2020}.

In practice, \cite{wu_adversarial_2020} found that one step of search is enough for finding approximately optimal weight perturbation. $\bm{v}$ can therefore be simplified (let $\rho = \gamma$) as:
\begin{align}
    \bm{v} &= \Pi_{\gamma}(\gamma \frac{\nabla_{\bm{v}} \mathbb{E}_{i \in M}[\mathcal{L}(\bm{x}_i+\bm{\delta}_i;\bm{\theta}+\bm{v})]}{\lVert \nabla_{\bm{v}} \mathbb{E}_{i \in M}[\mathcal{L}(\bm{x}_i+\bm{\delta}_i;\bm{\theta}+\bm{v})] \rVert} \lVert \bm{\theta} \rVert) \nonumber \\ 
    &= \gamma \frac{\nabla_{\bm{v}} \mathbb{E}_{i \in M}[\mathcal{L}(\bm{x}_i+\bm{\delta}_i;\bm{\theta}+\bm{v})]}{\lVert \nabla_{\bm{v}} \mathbb{E}_{i \in M}[\mathcal{L}(\bm{x}_i+\bm{\delta}_i;\bm{\theta}+\bm{v})] \rVert} \lVert \bm{\theta} \rVert
\label{equ: v one step update}
\end{align}
% As in the original work on AWP \cite{wu_adversarial_2020}, $\bm{v}$ is optimized with only one step of projected gradient descent so that the update rule can be simplified (see \cref{sec: awp review} for the derivation) to:
% \begin{equation}
%     \bm{v} = \gamma \frac{\nabla_{\bm{\theta}} \mathbb{E}_{i \in M}[\mathcal{L}(\bm{x}_i+\bm{\delta}_i;\bm{\theta})]}{\lVert \nabla_{\bm{\theta}} \mathbb{E}_{i \in M}[\mathcal{L}(\bm{x}_i+\bm{\delta}_i;\bm{\theta})] \rVert} \lVert \bm{\theta} \rVert
% \label{equ: v one step update}
% \end{equation}
%This extension is necessary to integrate weight loss smoothing into our framework, as described in detail in the following section.
\begin{algorithm}[t]
\SetAlgoCaptionLayout{small}
\SetAlgoCaptionSeparator{.}
\SetAlgoLined
\DontPrintSemicolon
\For{each batch}{
    \tcp{$m$ is batch size}
    \For{$i=1$ \KwTo $m$}{
        \tcp{adversarial input perturbation}
        $\bm{\delta}_i = \text{attack}(\bm{x}_i, y_i)$\;
        \tcp{adversarial vulnerability}
        $s_i = \text{vulnerability}(\bm{x}_i, \bm{\delta}_i)$\;    
    }
    \tcp{weight by vulnerability}
    $w_1, ..., w_m = \text{weight}(s_1, ..., s_m)$\;
    \tcp{enhance robustness by weights}
    $L = \sum_i^m (\mathcal{L}(\bm{x}_i + \bm{\delta}_i) + \text{robustify}(\bm{x}_i, w_i)) / m$\;
    \tcp{update model parameters}
    $\theta = \theta - l\nabla_{\theta}L$\;
}
\caption{The pseudo-code for the proposed general AT framework: Instance Adaptive Robustness Enhancement. }
\label{alg: general framework}
\end{algorithm}

Next, we consider how to weight instances according to their AV. We decided that regularization strength should depend on the relative order, instead of absolute value, of vulnerability so that the overall strength of regularization remains constant throughout training, even if the overall AV declines at the later stage of training. This is important for balancing the influence of AT and the additional robustifying methods. The regularization weight is generated linearly, based on the ranking of vulnerability within the batch, as follows:
\begin{equation}
    w_i = 1- \frac{r(\text{AV}_i)}{m} %= 1- \frac{r(\mathcal{L}(\bm{x}_i+\bm{\delta}_i; \bm{\theta}+\bm{v}) - \mathcal{L}(\bm{x}_i;\bm{\theta}))}{m}
\label{equ: linear weight scheme}
\end{equation}
where $r(\cdot)$ computes the ranking (indexed from 0 for the highest vulnerability) within the batch. Hence, the weights range from 1 (for the most vulnerable sample) to $\frac{1}{m}$ (for the least vulnerable).
This linear scheme is selected due to its simplicity and superiority over other options in performance (see \cref{sec: ablation study} for an empirical comparison with some alternatives).

Alternative robustifying methods that support instance-adaptive strength include  adversarial perturbation customization \cite{jia_-at_2022, yang_one_2022, cheng_cat_2022}, direct weight on loss \cite{zhang_geometry-aware_2021} and loss smoothness regularization \cite{li_understanding_2023}.
Typically, adversarial perturbation customization modifies the configuration of adversarial generation for each sample to reflect the required regularization strength, e.g., a large perturbation budget, $\epsilon$, corresponding to a large strength. 
Applying this strategy naively in our framework is expected to double the computational overhead since adversarial examples will be first generated to measure AV and then re-generated using the modified configuration that is customized to the vulnerability of each sample. The increased computational burden can be very costly when the training adversary is multi-step. In contrast,  
the direct weight on loss method weights each instance directly via a separate coefficient on the overall loss. It adds virtually no extra computational cost, but was observed before by \cite{jia_-at_2022, wang_self-ensemble_2022} to induce the model to overfit the training adversary resulting in a false security. Therefore, direct weight on loss is computationally efficient but ineffective.
Different from the above two methods, loss smoothness regularization has been widely verified to be effective in improving AT \cite{wang_improving_2020, wu_adversarial_2020} and can be computationally efficient if implemented properly \cite{li_understanding_2023}. Thus, we adopt loss smoothness regularization as  the robustifying method to realize our framework.

\begin{algorithm}[t]
\SetAlgoCaptionLayout{small}
\SetAlgoCaptionSeparator{.}
\SetAlgoLined
\DontPrintSemicolon
\For{each batch}{
    \tcp{$m$ is batch size}
    \For{$i=1$ \KwTo $m$}{
        \tcp{adversarial input perturbation}
        $\bm{\delta}_i = \text{PGD}(\bm{x}_i, y_i)$\;
    }
    \tcp{adversarial weight perturbation}
    $\bm{v} = \gamma \frac{\nabla_{\bm{\theta}} \mathbb{E}_{i \in M} [\mathcal{L}(\bm{x}_i+\bm{\delta}_i;\bm{\theta})]}{\lVert \nabla_{\bm{\theta}} \mathbb{E}_{i \in M}[\mathcal{L}(\bm{x}_i+\bm{\delta}_i;\bm{\theta})] \rVert} \lVert \bm{\theta} \rVert$\;
    \For{$i=1$ \KwTo $m$}{
        \tcp{weight by vulnerability}
        $w_i = 1- r(\mathcal{L}(\bm{x}_i+\bm{\delta}_i; \bm{\theta}+\bm{v}) - \mathcal{L}(\bm{x}_i;\bm{\theta}))/m$ \;
    }
    \tcp{input and weight smoothing}
    \For{$i=1$ \KwTo $m$}{
        $o_i = \lVert f(\bm{x}_i+\bm{\delta}_i;\bm{\theta}+\bm{v}) - f(\bm{x}_i; \bm{\theta}) \rVert_2^2$\;
    }
    $L = \sum_i^m (\mathcal{L}(\bm{x}_i+\bm{\delta}_i;\bm{\theta}+\bm{v}) + \lambda \cdot w_i \cdot o_i) / m$\;
    \tcp{update model parameters}
    $\theta = \theta - l\nabla_{\theta}L$\;
}
\caption{The pseudo-code for the proposed specific AT method: Instance adaptive Smoothness Enhanced Adversarial Training.}
\label{alg: instance-wise smoothness enhanced adversarial training}
\end{algorithm}
% AV measures theoretically the worst-case smoothness so it is supposed to be most sensitive to the variation of loss. 

\subsection{Jointly Smoothing Input and Weight Loss Surfaces} \label{sec: jointly smoothing input and weight loss}

To robustify the model in addition to AT, we propose a new regularization method that jointly smooths both input and weight loss landscapes. The idea of joint smoothing is motivated by the observation in \cref{sec: prevalence across various robust training schemes} that input and weight loss smoothing improve AT in a complementary way.
The proposed regularizer enforces prediction Logit Stability against both adversarial Input and Weight perturbation (LSIW) so that the model's predicted logits remains, ideally, constant when the input and the weights are both adversarially perturbed. Specifically, we penalize the logit variation raised by input perturbation, $\bm{\delta}$, and weight perturbation, $\bm{v}$, as
\begin{equation}
    \mathbb{E}_{i \in M} \lVert f(\bm{x}_i+\bm{\delta}_i; \bm{\theta}+\bm{v}) - f(\bm{x}_i;\bm{\theta}) \rVert_2^2
\label{equ: lsiw regularizer}
\end{equation}

Loss smoothness can be regularized in principle through logits and gradients.
Gradient regularization constrains the loss gradient instead of the predicted logits. However, this requires double-backpropagation which is computationally expensive. In contrast, logit regularization adds only a marginal expense for computing the regularization loss. Therefore, logit regularization is much more computationally efficient than gradient regularization. Moreover, logit regularization empirically outperforms gradient regularization in terms of robustness improvement and the trade-off between accuracy and robustness \cite{li_understanding_2023}. Therefore, we adopt logit regularization to smooth loss.

We acknowledge that the idea of jointly smoothing input and weight loss was explored before in \cite{wu_adversarial_2020}, but to our best knowledge, stabilizing predicted logits against adversarial weight perturbation is novel.
The previous work combined adversarial weight perturbation with input loss smoothing using the method (named TRADE-AWP in the original work):
\begin{equation}
    \text{KL}(f(\bm{x};\bm{\theta}+\bm{v}), f(\bm{x}+\bm{\delta}; \bm{\theta}+\bm{v}))
\label{equ: trade-awp}
\end{equation}
Hence, in contrast to our approach in this previous work both clean and adversarial examples were computed using the perturbed model, i.e., $\bm{\theta}+\bm{v}$. We argue that adversarial weight perturbation is not fully utilized in this paradigm since the logit variation caused by weight perturbation is not explicitly constrained by the outer Kullback-Leibler (KL) divergence. Theoretically, a stronger regularization can be realized by forcing the predicted logits to be same between clean examples on the unperturbed model and adversarial examples on the perturbed model, as in \cref{equ: lsiw regularizer}. The performance of these two approaches is compared in \cref{sec: ablation study}. These empirical results confirm the superiority of our approach over the previous one. Another difference between \cref{equ: lsiw regularizer} and \cref{equ: trade-awp} is the metric used to measure the similarity or distance between two prediction logits. Squared $\ell_2$-norm is adopted in our solution due to its superior performance as evaluated in \cref{sec: ablation study}.

\subsection{Optimization} \label{sec: optimization}
Finally, we combine AWP-based AT with the proposed weight scheme and regularization method to get the overall training loss:
\begin{equation}
    \mathbb{E}_{i \in M}[\mathcal{L}(\bm{x}_i+\bm{\delta}_i; \bm{\theta}+\bm{v}) + \lambda w_i\lVert f(\bm{x}_i+\bm{\delta}_i; \bm{\theta}+\bm{v}) - f(\bm{x}_i;\bm{\theta}) \rVert_2^2]
    \label{equ: overall loss}
\end{equation}
There are two hyper-parameters, $\lambda$ and $\gamma$, in our method. $\lambda$ controls the strength of joint regularizer. $\gamma$ in \cref{equ: v one step update} directly controls the strength of adversarial weight perturbation and also implicitly affect the strength of joint regularizer.

In practice, modern machine learning frameworks \cite{paszke_pytorch_2019} cannot directly compute the gradients of \cref{equ: overall loss} w.r.t. $\bm{\theta}$ in one backward pass on one model because the model used to compute $f(\bm{x}_i;\bm{\theta})$ will be altered by adversarial weight perturbation before backpropagation. To derive the update rule for gradient descent, we first rewrite \cref{equ: overall loss} as a function of two models parameterized by $\bm{\theta}'=\bm{\theta}+\bm{v}$ and $\bm{\theta}$ separately:
\begin{equation}
    L(f(\bm{x}+\bm{\delta}; \bm{\theta}'), f(\bm{x};\bm{\theta}))
\label{equ: two models loss formulation}
\end{equation}
Next, we apply the Chain rule to separate the gradient of \cref{equ: two models loss formulation} w.r.t. $\bm{\theta}$ into the sum of two individual backward passes:
\begin{align}
    \frac{\partial L}{\partial \bm{\theta}} =& \frac{\partial L}{\partial f(\bm{x}+\bm{\delta}; \bm{\theta}')}\frac{\partial f(\bm{x}+\bm{\delta}; \bm{\theta}')}{\partial \bm{\theta}'}\frac{\partial \bm{\theta}'}{\partial \bm{\theta}} \nonumber \\
    &+ \frac{\partial L}{\partial f(\bm{x}; \bm{\theta})}\frac{\partial f(\bm{x}; \bm{\theta})}{\partial \bm{\theta}}
\label{equ: gradient of overall loss}
\end{align}

After obtaining the gradients, we update the model's parameters following the method used for AWP \cite{wu_adversarial_2020} as:
\begin{equation}
    \bm{\theta} \leftarrow (\bm{\theta} + \bm{v}) - l \cdot {\cref{equ: gradient of overall loss}} - \bm{v}
\end{equation}
$l$ is the learning rate.

\subsection{Efficiency Analysis} \label{sec: efficiency analysis}
%We analyze here the computational efficiency of our method. 
The computational cost of the proposed method, ISEAT, is mainly composed of three components: AT, adversarial weight perturbation and logit stability regularization. Both AT and adversarial weight perturbation involve an inner maximization process using PGD, so their cost increases linearly with the number of iterations used for the inner optimization. By default, we use 10 and 1 iterations, respectively, for determining the input and weight perturbations. This is in accordance with common practice. Our implementation of logit stability regularization, LSIW, in practice adds one more forward and backward pass for $f(\bm{x};\bm{\theta})$ as required by \cref{equ: gradient of overall loss}. The time consumption is assessed empirically in \cref{sec-computational_efficiency}.

\section{Results} \label{sec: result}

The experiments in this section were based on the following setup unless otherwise specified. We evaluated our method with model architectures Wide ResNet34-10 (WRN34-10) \cite{zagoruyko_wide_2016} on dataset CIFAR10 \cite{krizhevsky_learning_2009} and PreAct ResNet18 (PRN18) \cite{he_identity_2016} on datasets CIFAR100 \cite{krizhevsky_learning_2009} and SVHN \cite{netzer_reading_2011}.
Models were trained by stochastic gradient descent for 200 epochs with an initial learning rate 0.1 for CIFAR10/100 and 0.01 for SVHN, divided by 10 at 50\% and 75\% of epochs. The momentum was 0.9, the weight decay was 5e-4 and the batch size was 128. The default data augmentation for CIFAR10/100 was horizontal flip (applied at half chance) and random crop (with 4 pixel padding). No data augmentation was applied to SVHN. Experiments were run on Tesla V100, A100 and RTX 3080Ti GPUs. All results generated by us were averaged over 3 runs.

For AT, we used $\ell_{\infty}$ projected gradient descent attack \cite{madry_towards_2018} with a perturbation budget, $\epsilon$, of 8/255. The number of steps was 10 and the step size was 2/255 for CIFAR10/100 and 1/255 for SVHN.
To stabilize the training on SVHN, the perturbation budget, $\epsilon$, was increased from 0 to 8/255 linearly in the first five epochs and then kept constant for the remaining epochs, as suggested by \cite{andriushchenko_understanding_2020}. Robustness was evaluated against AutoAttack \cite{croce_reliable_2020} using the implementation of \cite{kim_torchattacks_2021}. Note that, following \cite{rice_overfitting_2020}, we tracked PGD10 robustness on the test set at the end of each epoch during training and selected the checkpoint with the highest PGD10 robustness, i.e., the "best" checkpoint to report robustness. 

We compare our method with related works including AWP \cite{wu_adversarial_2020}, TRADE \cite{zhang_theoretically_2019}, InfoAT \cite{xu_infoat_2022}, RWP \cite{yu_robust_2022}, GAIRAT \cite{zhang_geometry-aware_2021}, FAT \cite{zhang_attacks_2020}, MART \cite{wang_improving_2020} and LAS-AT \cite{jia_-at_2022} on CIFAR10. All their result were copied from the original work or other published source like RobustBench \cite{croce_robustbench_2021}. They used the same model architecture and the same, or very similar, training settings as we did. We additionally evaluated the performance of our method when combined with the data augmentation method IDBH (weak variant) \cite{li_data_2023} and with extra data like RST \cite{carmon_unlabeled_2019} to benchmark state-of-the-art robustness. We observed that our method when combined with IDBH, akin to \cite{rebuffi_data_2021}, benefited from training longer so the total number of training epochs was increased to 400. Note that AT alone with IDBH (IBDH+AT) degenerated as the length of training was increased, so we report its performance with the default settings. For experiments with extra data, we used WideResNet28-10 instead of WideResNet34-10 to align with experimental protocols used in related works for a fair comparison. We adopted the same extra data as \cite{carmon_unlabeled_2019}, i.e., 500K unlabled data from dataset 80 Million TinyImages (80M-TI) with pseudo-labels\footnote{The extra data was downloaded from the official git repository of \cite{carmon_unlabeled_2019}: \url{https://github.com/yaircarmon/semisup-adv}.}. 
As in \cite{carmon_unlabeled_2019}, extra data was included in the ratio 1:1 with the CIFAR10 data in each training mini-batch, so the effective batch size became 256.
%and the number of samples seen by the model during training was equivalent to those that would have been used in 400 epochs without additional data.
%effective training length was 400 epochs.

The hyper-parameters of our method were optimized using grid search. The optimal values found were: $\lambda=0.1$ and $\gamma=0.007$ for CIFAR10; $\lambda=0.1$ and $\gamma=0.005$ for CIFAR10 with IDBH; $\lambda=0.01$ and $\gamma=0.005$ for CIFAR10 with extra data; $\lambda=0.1$ and $\gamma=0.005$ for CIFAR100; $\lambda=0.1$ and $\gamma=0.009$ for SVHN. We observed that jointly smoothing input and weight loss with a large learning rate (0.1 in this case) degraded both accuracy and robustness due to over-regularization. Therefore, we adopted a warm-up strategy for $\lambda$ on CIFAR10/100: $\lambda$ was set to 0 during the initial epochs when the learning rate was large, and $\lambda$ was set to the optimal value after first decay of the learning rate. Note that this strategy was not applied to the experiments with SVHN because the initial learning rate on SVHN was already small.

\subsection{Benchmarking State-of-the-Art Robustness} \label{sec: benchmarking robustness}
% \usepackage{booktabs}
% \usepackage{graphicx}
% \usepackage[normalem]{ulem}
\begin{table}[tbp]
\centering
\caption{Test accuracy and robustness for networks trained on CIFAR10 using our method and related methods. Results above the double line are for WRN34-10 without extra data and results below the double line are for WRN28-10 with extra data. The best result is highlighted for each metric in each block. The standard deviation is indicated by the value after the $\pm$ sign if evaluated by us or reported in the original work, otherwise omitted from the table.}
\label{tab: cifar10 comparison}
%\resizebox{\columnwidth}{!}{%
\begin{tabular}{@{}lccll@{}}
\toprule
\multirow{2}{*}{Method} & \multirow{2}{*}{Model} & Extra & Accuracy & Robustness \\ 
 & & Data &  (\%) &  (\%)\\\midrule
AT         & \multirow{15}{*}{\rotatebox[origin=c]{90}{------------------ WRN34-10 ------------------}} & -      & 85.90 $\pm$ 0.57              & 53.42 $\pm$ 0.59               \\
AT-AWP     & & -      & 85.57 $\pm$ 0.40              & 54.04 $\pm$ 0.40               \\
TRADE      & & -      & 85.72                & 53.40                \\
InfoAT     & & -      & 85.62                & 52.86                \\
GAIRAT     & & -      & 86.30                & 40.30                \\
FAT        & & -      & {\ul \textbf{87.97}}            & 47.48                \\
RWP        & & -      & 86.86 $\pm$ 0.51         & 54.61  $\pm$ 0.11        \\
MART       & & -      & 84.17 $\pm$ 0.40               & 51.10  $\pm$ 0.40              \\
MART-AWP   & & -      & 84.43 $\pm$ 0.40               & 54.23  $\pm$ 0.40              \\
LAS-AT     & & -      & 86.23                & 53.58                \\
LAS-AWP    & & -      & 87.74 & 55.52                \\
ISEAT (ours) &  & -      & 86.02    $\pm$  0.36          & 56.54 $\pm$ 0.36\\
ISEAT (ours)+SWA\hspace*{-3mm}  & & -      & 85.95 $\pm$ 0.09              & {\ul \textbf{57.09}} $\pm$ 0.13 \\ \cmidrule{1-1}\cmidrule{3-5}
IDBH+AT    & & -      & 87.03  $\pm$  1.58        & 54.16 $\pm$  0.70          \\
% IDBH+AT-AWP    & & -      & 88.47  $\pm$  0.45            & 57.88 $\pm$ 0.23              \\
IDBH+ISEAT (ours)\hspace*{-3mm}  & & -      & {\ul \textbf{88.50}} $\pm$ 0.11 & {\ul \textbf{59.32}} $\pm$ 0.08 \\
\midrule\midrule
RST        & \multirow{7}{*}{\rotatebox[origin=c]{90}{----- WRN28-10 -----}} & \multirow{6}{*}{\rotatebox[origin=c]{90}{---- 80M-TI ----}} & 89.69  $\pm$ 0.40              & 59.53  $\pm$ 0.40              \\
RST+MART   & & & 87.50                & 56.29                \\
RST+GAIRAT & & & 89.36                & 59.64                \\
RST+AWP    & & & 88.25  $\pm$ 0.40              & 60.05  $\pm$ 0.40              \\
RST+RWP    & & & 88.87 $\pm$ 0.55           & 60.36   $\pm$  0.06   \\
ISEAT (ours) & & & {\ul \textbf{90.59}} $\pm$ 0.19 & {\ul \textbf{61.55}} $\pm$ 0.10 \\ 
\cmidrule{1-1}\cmidrule{3-5}
IDBH+ISEAT (ours)\hspace*{-3mm}  & & -      & 87.91 $\pm$ 0.18 & 58.55 $\pm$ 0.14 \\ 
\bottomrule
\end{tabular}%
%}
\end{table}\tabcolsep\tabcolseporig
As can be seen fro the results in \cref{tab: cifar10 comparison}, our method significantly improves both accuracy and robustness over the baseline in all evaluated settings. Specifically, it boosts robustness by $+3.12\%$ compared to AT in the default setting, by $+5.16\%$ when IDBH data augmentation is used, and by $+2.02\%$ compared to RST when extra real data from 80M-TI is used. More importantly, our method boosts accuracy as well suggesting a better trade-off between accuracy and robustness. 
By combining with IDBH, our method achieves a robustness of 58.55\% for WRN28-10 which is competitive with the baseline robustness of 59.53\% achieved by RST using additional real data. This significantly closes the gap between the robust performance of training with and without extra data. Finally, we highlight that our method achieves, to our best knowledge, state-of-the-art robustness $61.55\%$\footnote{A higher record of robustness, 62.80\%, was reported by \cite{gowal_uncovering_2021}. However, it is unfair to directly compare our result with theirs since we used significantly different training settings. \cite{gowal_uncovering_2021} replaced the ReLU activation function with SiLU in WRN28-10. They also changed the batch size from 128 to 512, the labeled-to-unlabeled ratio from 1:1 to 3:7 and the learning rate schedule from piecewise to cosine. All these modifications were observed to have an important effect on the robust performance according to their experimental results. We expect our method to achieve higher robustness when trained using customized settings, but our computational resources are insufficient to search for the optimal training setup: \cite{gowal_uncovering_2021} used 32 Google Cloud TPU v3 cores for training.} and $59.32\%$ for the settings with and without extra data on the corresponding model architectures respectively. 

Our method outperforms all existing instance-adaptive AT methods in terms of robustness. We compare our method with related works using the default setup and CIFAR10 (\cref{tab: cifar10 comparison}) since published results are available for this setup. Our method achieves the highest robustness, $56.54\%$, among all competitive works, which considerably exceeds the previous best record of $55.52\%$ achieved by LAS-AWP and the robustness of $54.23\%$ achieved by the most similar previous work, MART-AWP. Particularly, our method dramatically outperforms FAT by $+9.06\%$ in terms of robustness. FAT is one of the most recent contributions whose instance adaptation strategy contrasts ours, as described in \cref{sec: related works}. This supports our claim that this previous strategy for instance-adaptive AT is fundamentally defective. Furthermore, our method consistently achieves superior robustness compared to all available related works such as MART and GAIRAT in the condition with extra data.

Last, we find that the performance of our method can be further improved by $+0.55\%$ in the default setup when another weight smoothing technique Stochastic Weight Averaging (SWA) is integrated. However, we did not observe a similar performance boost in the other setting for CIFAR10. This suggests that our method may exhaust the benefits of weight smoothing in some settings, but not all.

% \usepackage{booktabs}
% \usepackage[normalem]{ulem}
\begin{table}[tbp]
\centering
\caption{Test accuracy and robustness of our method for PRN18 on CIFAR100 and SVHN.}
\label{tab: cifar100 svhn robustness}
\begin{tabular}{@{}llcc@{}}
\toprule
Dataset & Method & Accuracy (\%)             & Robustness (\%)           \\ \midrule
CIFAR100 & AT   & {\ul \textbf{56.15}} $\pm$ 1.15 & 25.12 $\pm$ 0.22                \\
CIFAR100 & ISEAT (ours) & 53.19 $\pm$ 0.23                & {\ul \textbf{28.17}} $\pm$ 0.14\\ \midrule
SVHN     & AT   & 90.55 $\pm$ 0.60                & 47.48 $\pm$ 0.59                \\
SVHN                        & ISEAT (ours)                       & {\ul \textbf{91.08}} $\pm$ 0.49 & {\ul \textbf{54.04}} $\pm$ 0.68\\ \bottomrule
\end{tabular}
\end{table}

\subsection{Generalization to Other Datasets}
%Our method is able to generalize across various datasets. 
Following common practice for testing generalization ability, we evaluate our method on the alternative datasets CIFAR100 and SVHN. As shown in \cref{tab: cifar100 svhn robustness}, our method significantly improves robustness over the baseline by $+3.05\%$ on CIFAR100 and by $+6.56\%$ on SVHN. It also slightly boosts accuracy on SVHN. Note that the magnitude of robustness improvement in a particular training setting generally depends on the degree of robust overfitting, which is connected to the unevenness of AV among training data. It is therefore reasonable for our method to perform differently on different datasets even using the same model architecture. Overall, the performance improvements across various datasets is consistent, which confirms that the proposed method is generally applicable.

\subsection{Ablation Study} \label{sec: ablation study}
We conducted ablation experiments to justify the design of our method and illuminate the mechanism behind its effectiveness. Experiments were performed using  WRN34-10 on CIFAR10. To ensure a fair comparison, the approaches were applied to fine-tune the same model. This base model had been previously trained using AT with the default training setup, as described in \cref{sec: result}. Fine-tuning was performed for 40 epochs. The initial learning rate was 0.01 and decayed to 0.001 after 20 epochs. 

We first assess the contribution of different components in our method. It can be observed in \cref{tab: component comparison} that both the components of our method, (adaptively weighted) input loss smoothing and weight loss smoothing, can individually improve robustness over the baseline to a great extent, $+1.08\%$ and $+1.84\%$ respectively. This confirms that they both play a vital role in our method. Furthermore, combining them together (the proposed method) achieves a greater robustness boost, $+3.04\%$, compared to either of them alone. This combined boost is greater than the arithmetic sum of the performance increases of the individual components ($3.04\% > 1.08\%+1.84\%$) suggesting that these two components are complementary to each other.

% \usepackage{booktabs}
% \usepackage[normalem]{ulem}
%\tabcolsep3.5pt
\begin{table}[]
\centering
\caption{The performance contribution of each component in the proposed ISEAT method. Time is an average measured for processing one
mini-batch on a Nvidia RTX 3080Ti in seconds.}
\label{tab: component comparison}
%\resizebox{\columnwidth}{!}{
\begin{tabular}{@{}lccl@{}}
\toprule
Method & Accuracy (\%)           & Robustness (\%)         & Time (s) \\ \midrule
AT                         & 85.90 $\pm$ 0.57                & 53.42 $\pm$ 0.59& {\ul \textbf{0.253}}                    \\
+input loss              & \multirow{2}{*}{84.10 $\pm$ 0.27} & \multirow{2}{*}{54.50 $\pm$ 0.17}           & \multirow{2}{*}{0.268 (+6\%) }            \\
\hspace*{1ex}smoothing\\
+weight loss             & \multirow{2}{*}{{\ul \textbf{86.04}} $\pm$ 0.27} & \multirow{2}{*}{55.26 $\pm$ 0.15} & \multirow{2}{*}{0.277 (+9\%) }  \\
\hspace*{1ex}smoothing\\
+both (ISEAT)              & 85.63 $\pm$ 0.13               & {\ul \textbf{56.46}} $\pm$ 0.14 & 0.298 (+18\%)            \\ \bottomrule
\end{tabular}
%}
\end{table}\tabcolsep\tabcolseporig

Next, we examine our design of input loss smoothness regularizier. We first verify the choice of distance metric used to measure the dissimilarity between two predicted logits. As shown in \cref{tab: logit stability reg ablation}, squared $\ell_2$-norm (the adopted method) performs slightly better than KL-divergence (used by MART \cite{wang_improving_2020}) in terms of both accuracy and robustness. Moreover, we compare the performance of the linear weight scheme (the chosen method) with the top-10\% weight scheme (used in the preliminary experiments reported in \cref{sec: connection to robust overfitting}, see \cref{equ: topn regularization}) and unweighted (or uniform) scheme. It can be observed in \cref{tab: logit stability reg ablation} that the weighted schemes, either linear or top-10\%, considerably improve both accuracy and robustness over the unweighted scheme, and among the weighted schemes, the linear one outperforms the top-10\% scheme regarding both accuracy and robustness. Overall, a linear weight scheme with squared $\ell_2$-norm is empirically the best among all evaluated solutions.

% \usepackage{booktabs}
% \usepackage[normalem]{ulem}
\begin{table}[]
\centering
\caption{The performance of logit stability regularization with different distance metrics and weight schemes. "Distance" denotes the metric used to measure the discrepancy between two predicted logits. "Weight" denotes the weight scheme.}
\label{tab: logit stability reg ablation}
%\resizebox{\columnwidth}{!}{
\begin{tabular}{@{}llcc@{}}
\toprule
Distance & Weight & Accuracy  (\%)           & Robustness (\%)         \\ \midrule
AT         &                          & {\ul \textbf{85.90}} $\pm$ 0.57 & 53.42 $\pm$ 0.59               \\
KL-divergence                & unweighted                 & 85.07 $\pm$ 0.31               & 56.08 $\pm$ 0.32               \\
Squared $\ell_2$-norm                    & unweighted                 & 85.15 $\pm$ 0.70               & 56.20 $\pm$ 0.19              \\
Squared $\ell_2$-norm                    & top-10\%                       & 85.53 $\pm$ 0.03               & 56.36 $\pm$ 0.02              \\
Squared $\ell_2$-norm                   & linear                     & 85.63 $\pm$ 0.13               & {\ul \textbf{56.46}} $\pm$ 0.14 \\ \bottomrule
\end{tabular}
%}
\end{table}\tabcolsep\tabcolseporig

Last, we examine the effectiveness of our approach to combining input and weight loss smoothing. We compare our proposal, LSIW, with Logit Stability regularization against Input perturbation only (LSI) and TRADE-AWP. The regularization loss of these methods is described in \cref{tab: weight smoothing ablation}. For more technical detail, please refer to \cref{sec: jointly smoothing input and weight loss}. We observe in \cref{tab: weight smoothing ablation} that our approach achieves significantly higher accuracy and robustness than the others. This supports our hypothesis that stabilizing logits against both input and weight adversarial perturbation makes a better use of adversarial weight perturbation, and hence, results in a more effective smoothness regularization.

% \usepackage{booktabs}
% \usepackage{graphicx}
% \usepackage[normalem]{ulem}
\tabcolsep1.5pt
\begin{table}[tbp]
\centering
\caption{The performance of different approaches to jointly smoothing input and weight loss landscapes. $\bm{x}'$ and $\bm{\theta}'$ refer to the perturbed input and weight respectively. For a fair comparison, the original distance metric, KL-divergence, in TRADE-AWP (\cref{equ: trade-awp}) was replaced by squared $\ell_2$-norm to align with the other methods.}
\label{tab: weight smoothing ablation}
%\resizebox{\columnwidth}{!}{%
\begin{tabular}{@{}llcc@{}}
\toprule
Method & Smoothness Loss & Accuracy (\%)             & Robustness (\%)          \\ \midrule
AT                   &      & {\ul \textbf{85.90}} $\pm$ 0.57 & 53.42 $\pm$ 0.59                \\
+LSI                 &  $\lVert f(\bm{x}'; \bm{\theta}) - f(\bm{x}; \bm{\theta}) \rVert_2^2$   & 85.49 $\pm$ 0.50                & 55.38 $\pm$ 0.32              \\
+TRADE-AWP          &   $\lVert f(\bm{x}'; \bm{\theta}') - f(\bm{x}; \bm{\theta}') \rVert_2^2$   & 85.52 $\pm$ 0.29               & 55.82 $\pm$ 0.46               \\
+LSIW (ours)        &  $\lVert f(\bm{x}'; \bm{\theta}') - f(\bm{x}; \bm{\theta}) \rVert_2^2$    & 85.63 $\pm$ 0.13              & {\ul \textbf{56.46}} $\pm$ 0.14 \\ \bottomrule
\end{tabular}%
%}
\end{table}\tabcolsep\tabcolseporig

\subsection{Computational Efficiency}
\label{sec-computational_efficiency}
%We assess here the computational cost of our method. 
It can be seen from the results in \cref{tab: component comparison} that smoothing input loss landscape alone (i.e., weighted logit stability regularization) adds about 6\% computational overhead, and smoothing weight loss landscape alone (i.e., AWP) adds around 9\% computational overhead compared to AT. Jointly smoothing both input and weight loss landscapes using the proposed ISEAT method introduces an overhead of approximately 18\% compared to AT. The extra cost of our method is greater than the sum of the extra cost of two separate smoothing components ($18\%>6\%+9\%$) because it requires additional forward and backward passes to compute the gradient of the proposed regularization. For more detail, please refer to \cref{sec: optimization} and \cref{sec: efficiency analysis}.

% \subsection{how the proposal works}
% the distribution of margin

% excessive margin remains. this can be a future work.

\section{Conclusion}
This work investigated how adversarial vulnerability evolves during AT from an instance-wise perspective. We observed the model was trained to be more robust for some samples and, meanwhile, more vulnerable at some others resulting in an increasingly uneven distribution of adversarial vulnerability among training data.
%we found that the model became disordered robust, i.e., excessive margin and benign adversarial direction at a considerable amount of training samples. 
We theoretically proposed an alternative optimization path to minimize adversarial loss as an account for this phenomenon. Motivated by the above observations, we first proposed a new AT framework that enhances robustness at each sample with strength adapted to its adversarial vulnerability. We then realized it with a novel regularization method that jointly smooths input and weight loss landscapes. Our proposed method is novel in a number of respects: 1) adapting regularization to instance-wise adversarial vulnerability is new and contrasts the popular existing  strategy; 2) stabilizing logit against adversarial input and weight perturbation simultaneously is novel and more effective than the previous approaches. Experimental result shows our method outperforms all related works and significantly improves robustness w.r.t. the AT baseline. Extensive ablation studies demonstrate the vital contribution of the proposed instance adaptation strategy and smoothness regularizer in our method.

In addition to finding that AT results in an uneven distribution of adversarial vulnerability among training data, we also observed that for a considerable proportion of samples the model was excessively robust, such that even very large perturbations, making the sample unrecognizable to a human, failed to influence the prediction made by the network. 
One limitation of this work is that the proposed method, albeit effective in improving robustness, does not mitigate the issue of "disordered robustness". Future work might usefully explore this problem to further improve the performance of AT. A better trade-off between accuracy and robustness is anticipated if disordered robustness is alleviated.

\ifCLASSOPTIONcompsoc
  \section*{Acknowledgments}
\else
  \section*{Acknowledgment}
\fi

The authors acknowledge the use of the research computing facility at King’s College London, King's Computational Research, Engineering and Technology Environment (CREATE), and the Joint Academic Data science Endeavour (JADE) facility. This research was funded by the King's - China Scholarship Council (K-CSC).

\ifCLASSOPTIONcaptionsoff
  \newpage
\fi



\bibliographystyle{IEEEtran}
\bibliography{references}


% \end{thebibliography}


\begin{IEEEbiography}[{\includegraphics[width=1in,height=1.25in,clip,keepaspectratio]{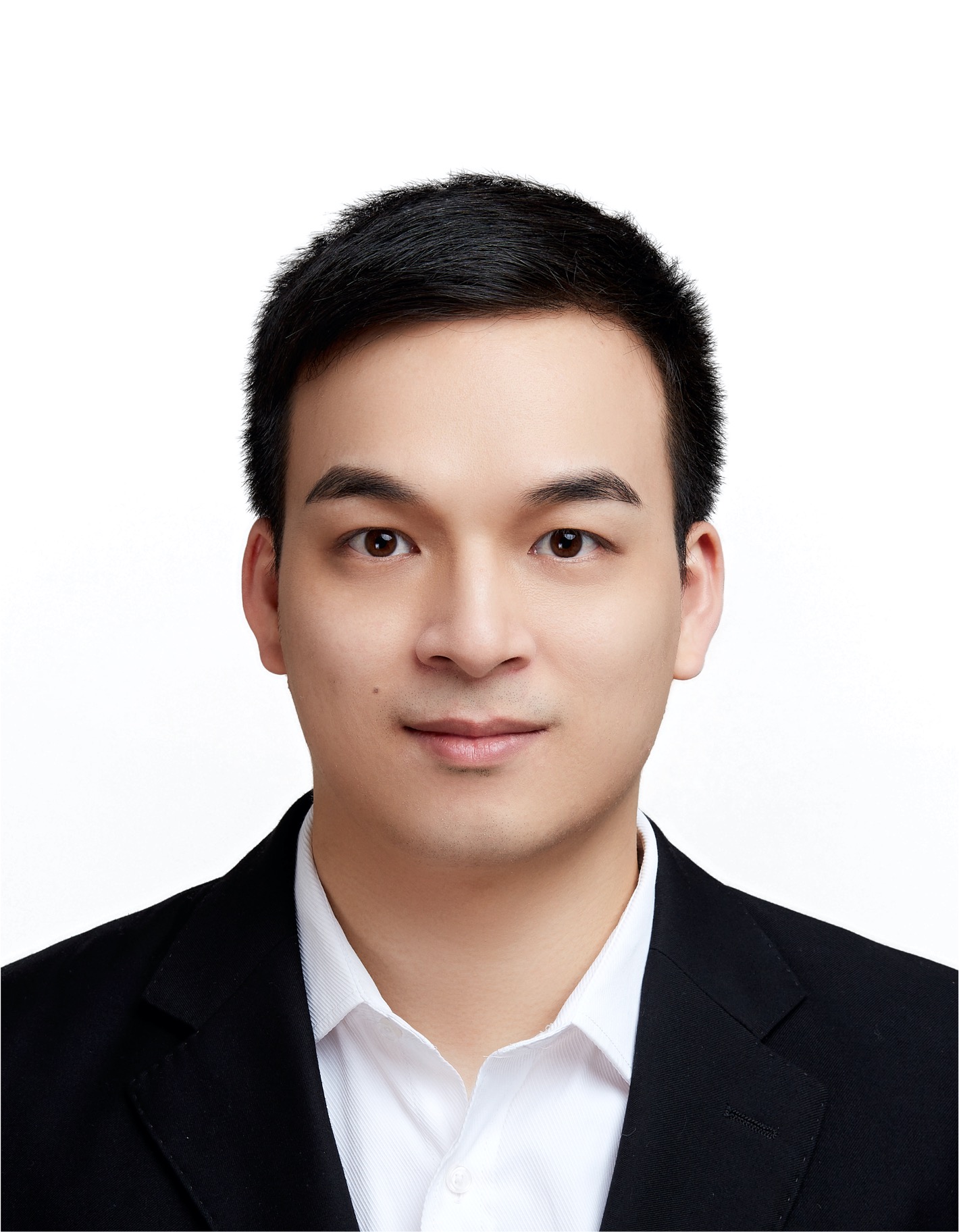}}]{Lin Li}
received a B.M. degree from Xiamen University and a M.Sc. degree in computing from Imperial College London. He is currently a Ph.D. student in computer science at the Department of Informatics, King’s College London. His research interest includes adversarial robustness and interpretability of deep learning.
\end{IEEEbiography}

\begin{IEEEbiography}[{\includegraphics[width=1in,height=1.25in,clip,keepaspectratio]{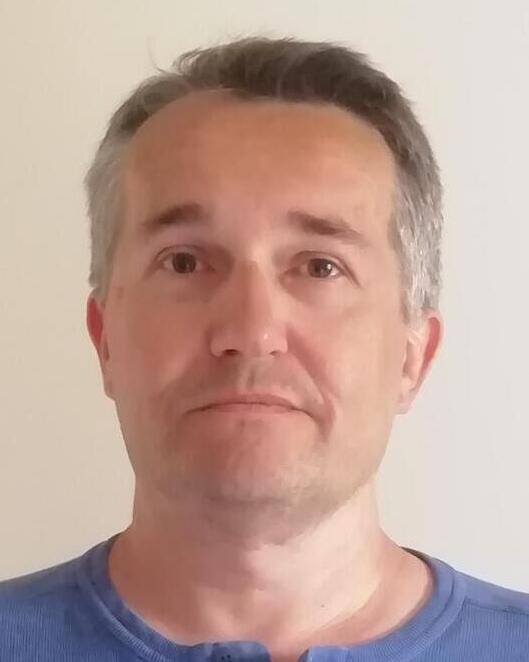}}]{Michael Spratling}
received a B.Eng. degree in engineering science from Loughborough University and M.Sc. and Ph.D. degrees in artificial intelligence and neural computation from the University of Edinburgh. He is currently Reader in Computational Neuroscience and Visual Cognition at the Department of Informatics, King’s College London. His research is concerned with understanding the computational and neural mechanisms underlying visual perception, and developing biologically-inspired neural networks to solve problems in computer vision and machine learning.
\end{IEEEbiography}
\vfill
%\newpage


%\vfill


\end{document}